\documentclass[ijds,nonblindrev]{informs-ijds}

\OneAndAHalfSpacedXI
\usepackage{natbib}
 \bibpunct[, ]{(}{)}{,}{a}{}{,}%
 \def\bibfont{\small}%

\usepackage{algorithm, multirow}
\usepackage[]{algpseudocode}
\usepackage{tikz}
\usepackage{underscore}
\usepackage[english]{babel}
\tikzstyle{mynode} = [circle, draw, fill=white,minimum size = 1.00cm]

\TheoremsNumberedThrough     % Preferred (Theorem 1, Lemma 1, Theorem 2)
\ECRepeatTheorems

\EquationsNumberedThrough    % Default: (1), (2), ...
\begin{document}
%%%%%%%%%%%%%%%%

%\setcounter{page}{1}
%\DOI{10.1287/xxxx.0000.0000}%


\RUNTITLE{Conjecturing-Based Discovery of Patterns in Data}

% \TITLE{Bundling Information Goods of Decreasing Value}
\TITLE{Conjecturing-Based Discovery of Patterns in Data\footnote{Data Ethics and Reproducibility Note: Code and data to reproduce results are available here: https://github.com/jpbrooks/conjecturing.  COVID-19 synthetic patient data was obtained as part of the Veterans Health Administration (VHA) Innovation Ecosystem and precisionFDA COVID-19 Risk Factor Modeling Challenge and is used here with permission from the Food and Drug Administration (FDA).}}

\ARTICLEAUTHORS{%
\AUTHOR{J. Paul Brooks}
\AFF{Department of Information Systems, Virginia Commonwealth University, Richmond, VA 23284, \EMAIL{jpbrooks@vcu.edu}} %, \URL{}}
\AUTHOR{David J. Edwards}
\AFF{Department of Statistical Sciences and Operations Research, Virginia Commonwealth University, Richmond, VA 23284, \EMAIL{dedwards7@vcu.edu}}
\AUTHOR{Craig E. Larson}
\AFF{Department of Mathematics and Applied Mathematics, Virginia Commonwealth University, Richmond, VA 23284, \EMAIL{clarson@vcu.edu}}
\AUTHOR{Nico Van Cleemput}
\AFF{Department of Applied Mathematics, Computer Science and Statistics, Ghent University, Ghent, Belgium, \EMAIL{nico.vancleemput@gmail.com}}
} % end of the block

\ABSTRACT{%
%Modern machine learning methods are designed to exploit complex patterns in data regardless of their form, while not necessarily revealing them to the investigator.
We propose the use of a conjecturing machine that {suggests} feature relationships in the form of bounds involving nonlinear terms for numerical features and boolean expressions for categorical features.  The proposed \textsc{Conjecturing} framework recovers known nonlinear and boolean relationships among features from data.   In both settings, true underlying relationships are revealed.  We then compare the method to a previously-proposed framework for symbolic regression { on the ability to} recover equations that are satisfied among features in a dataset. 
The framework is then applied to patient-level data regarding COVID-19 outcomes to suggest possible risk factors that are confirmed in the medical literature.  
}%

%\KEYWORDS{deterministic inventory theory; infinite linear programming duality;
%  existence of optimal policies; semi-Markov decision process; cyclic schedule}

   \HISTORY{} 

\maketitle


\section{Introduction}
Modern machine learning methods allow one to leverage complex relationships present in data to generate accurate predictions but do not reveal them to the investigator.  We propose an automated conjecturing framework for discovering nonlinear and boolean relationships among the features in a given dataset.  Our primary goal is discovery - to provide the investigator with a manageable number of suggested relationships to inspire future investigation for validation.

The nonlinear relationships are produced in the form of bounds.  Bounds are useful for scientific discovery from numeric data because they 1) suggest direct and indirect relationships among features, 2) suggest a functional form for the relationships, and 3) can subsequently be used as boolean features (e.g., is this bound satisfied by an observation?) for discovering more complex boolean relationships.  Whereas previous related approaches seek to find equations for numeric data, our \textsc{Conjecturing} method produces bounds for numeric data, boolean expressions for discrete data, and bounds and boolean expressions for mixed data.

%that can be used to enhance the prediction of a response while providing model transparency.  In many situations such as medical and financial decision making, knowing the basis for predictions is important for reasons including understanding causal relationships, establishing trust, and assigning liability \citep{Stoyanovich20}.  

%In situations where black box models provide accurate predictions, there could be easily-understandable feature relationships that guide system behavior.  Our goal is to develop a framework for revealing the least complex nonlinear and boolean relationships.  For example, consider analyzing measurements on the gravitational force between several pairs of masses at various distances apart.  Below we simulate such a system and demonstrate how one can discover that force is directly proportional to the product of the masses and inversely proportional to the square of the distance.  We also demonstrate how the method can be used in other contexts to make discoveries.  

% discuss computational complexity?

\citet{UdrescuTegmark20} proposed a system called \textsc{AI Feynman} that combines deep learning with methods for symbolic regression to recover nonlinear relationships in data.  Impressively, they recover over 100 equations of varying complexity from data.  In contrast to \textsc{AI Feynman}, our \textsc{Conjecturing} framework uses Fajtlowicz's Dalmatian heuristic \citep{Fajt95} to discover bounds rather than equations.  Further, our framework can also be applied to categorical data to discover boolean relationships among features and already-discovered bounds.  This work represents the first application of the Dalmatian heuristic to learning both nonlinear and boolean relationships from data.  The bounds and conditions produce interpretable yet complex relationships.

\section{Background and Previous Related Work}
\label{relatedwork}

In this section, we provide background on our  \textsc{Conjecturing} framework including examples of uses of bounds and sufficient conditions, a description of the core algorithm, and a survey of previous related work.

\subsection{Conjectured Bounds and Sufficient Conditions}
The algorithm we use to conjecture feature relationships is an adaptation of an algorithm that was originally designed to conjecture relationships for mathematical objects.  To illustrate the potential value of bounds and sufficient conditions, we describe two problems and relevant results from graph theory.  This paper extends these ideas regarding bounds and sufficient conditions to learning from data.

A graph is a collection of nodes $V$ and edges $E$ that are ordered pairs of nodes.  Consider the problem of finding bounds for the independence number of a graph\footnote{The independence number is the largest number of nodes in a graph no two of which are contained in an edge.  The definition of independence number is not important for this example, but only the fact that with every graph is associated a number called the ``independence number''.}.  It is well known that the linear programming (LP) relaxation of an appropriate integer program provides a upper bound on the independence number \citep{Schr03}.  The Lov\'asz $\vartheta$ number of a graph also provides an upper bound that is known to be no larger than the LP relaxation bound for any graph \citep{Lova79}.  Therefore, the Lov\'asz $\vartheta$ bound dominates the LP relaxation bound, and such relationships are commonly pursued.  However, relationships among bounds can be more nuanced.  Consider a third bound on the independence number due to \citet{Haem79}.  For some graphs it is a stronger bound than Lov\'asz $\vartheta$ while on other graphs it is a weaker bound; for some graphs, Lov\'asz $\vartheta$ is a sharp bound and Haemers's bound is not while for other graphs, Haemers's bound is a sharp bound and Lov\'asz $\vartheta$ is not.  It remains an open question whether there are a ``small'' number of bounds where the largest for value for any graph would provide a sharp bound on the independence number.  In this paper, we describe a computational approach to discover bounds among numeric features in a dataset.  As with the independence number, collections of bounds can provide valuable insight into relationships for the system from which the data was collected.  

Now consider the problem of finding sufficient conditions for a graph to be Hamiltonian\footnote{A Hamiltonian graph is a graph with a spanning cycle \citep{West01}.  The definition of Hamiltonian is not important for this example, but only the fact that any graph either is or is not Hamiltonian.}.  \citet{chvatal:72} proved that for a graph $G$ with certain conditions on the vertex degrees, $G$ is Hamiltonian.  Also, \citet{ChvaErdo72} proved that if a graph satisfies a connectivity condition, then it is Hamiltonian.  These are two conditions that are sufficient for a graph to be Hamiltonian, but neither implies the other.  Some graphs satisfy both conditions, some graphs satisfy one condition, and some graphs satisfy neither condition.  The existence and discovery of a (small) set of sufficient conditions that characterize all Hamiltonian graphs remains an open area of research.  The pursuit of sufficient conditions of graph properties such as Hamiltonicity mirrors that of  bounds \citep{LarsVanc17}.  In the context of learning from data, we show how categorical data, together with bounds discovered among numeric features, can be used as input to a computational approach for generating sufficient conditions for a property of interest. 

\subsection{The Dalmatian Heuristic}
\label{dalmatiansec}
Our \textsc{Conjecturing} framework is based on an implementation of Fajtlowicz's Dalmatian heuristic \citep{Fajt95, LarsVanc17}.  The heuristic was originally implemented in {\sc Graffiti} \citep{Fajt95} which was the first program to produce research conjectures that led to new mathematical theory.  The program produces statements that are relations between mathematical {\it invariants} which are numerical attributes of examples.  Recent implementations of the Dalmatian heuristic have been applied to the discovery of relationships for graphs \citep{LarsVanc16} and game strategies \citep{BradEtal17}.  The heuristic was adapted to work with {\it properties} which are boolean attributes of examples by \citet{LarsVanc17}.   We built our framework using a more recent implementation of the Dalmatian heuristic available here: \url{http://nvcleemp.github.io/conjecturing/}.    
We now describe invariant conjecturing using Fajtolwicz's Dalmatian heuristic.  The inputs include the following.  Let $E$ be a set of examples of a given type (e.g., graphs or data observations). Let $A = \{\alpha_1, \alpha_2, \ldots, \alpha_m\}$ be real number invariants.  In this work the examples are $n$ data observations and the invariants are $m$ numeric features.  The real-numbered value of example $i$ for invariant $\alpha_j$ is $\alpha_j(i) = x_{ij}$ for $i=1,\ldots, n$ and $j=1,\ldots,m$.  Let $O$ be a collection of {\it unary operators} and {\it binary operators}.  Examples of unary operators include adding 1, squaring, square-rooting, and division by 2. Binary operators include addition, multiplication, and subtraction.  Let $\alpha^* \in A$ be the invariant for which upper and lower bounds are of interest, and let $\alpha^*(i)$ be the value of the invariant of interest for example $i$.
 
%We form a stream of algebraic functions of the invariants: $\alpha_1 + \alpha_2$, $\sqrt{\alpha_1}$, $\alpha_1\alpha_3$, $(\alpha_2 +\alpha_4)^2$, etc. 
% These expressions can then be used to form conjectured bounds for $\alpha^*$. If we are interested in upper bounds for $\alpha^*$, we can form the inequalities $\alpha^*\leq\alpha_1 + \alpha_2$, $\alpha^*\leq\sqrt{\alpha_1}$, $\alpha^*\leq\alpha_1\alpha_3$, $\alpha^*\leq(\alpha_2 +\alpha_4)^2$, etc. 

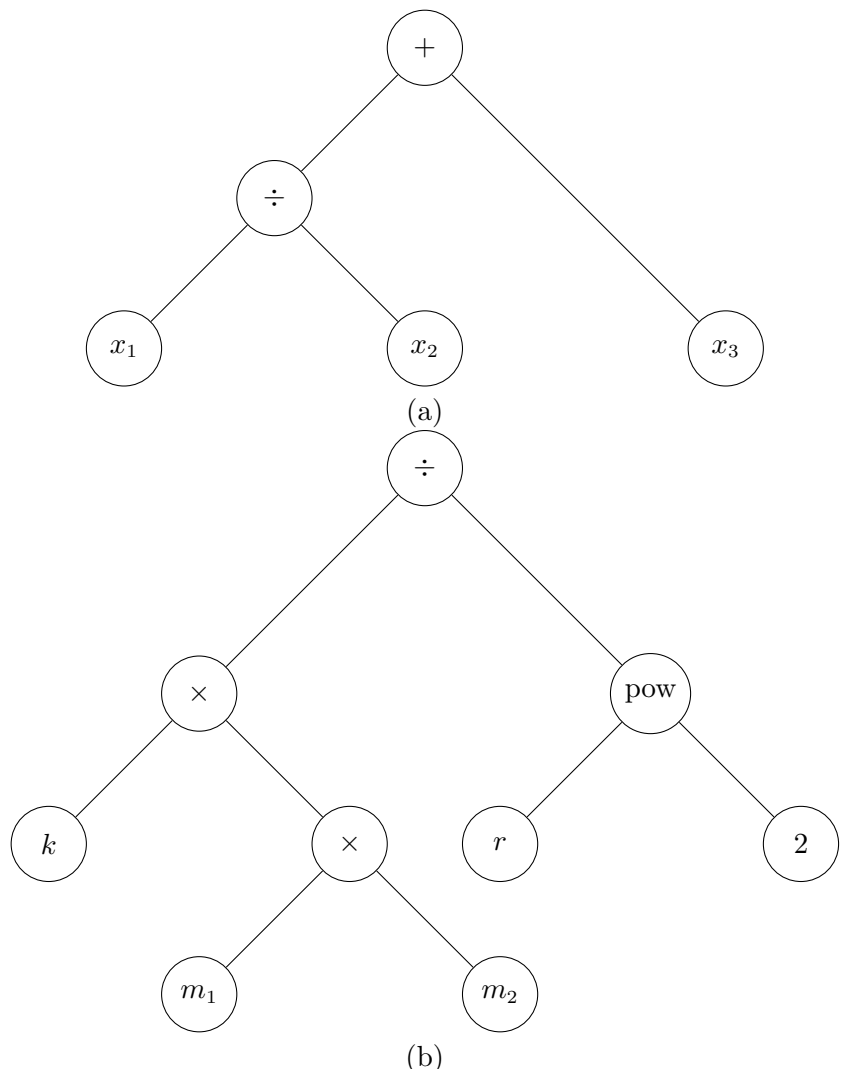
\begin{figure}
\begin{tabular}{c}
\begin{tikzpicture}[node distance=2.0cm]
 \node[mynode](top) {$+$};
 \node[mynode](div) [left of=top, below of=top] {$\div$};
 \node[mynode](x1) [left of=div, below of=div] {$x_1$};
 \node[mynode](x2) [right of=div, below of=div] {$x_2$};
 \node[mynode](x3) [right of=x2, node distance=4.0cm] {$x_3$};
 \draw[black] (top) to (div);
 \draw[black] (top) to (x3);
 \draw[black] (div) to (x1);
 \draw[black] (div) to (x2);
\end{tikzpicture}
\\
(a)\\
\begin{tikzpicture}[node distance=2.0cm]
 \node[mynode](top) {$\div$};
 \node[mynode](times) [left of=top, below of=top, node distance = 3.0cm] {$\times$};
 \node[mynode](pow) [right of=top, below of=top, node distance = 3.0cm] {pow};
 \node[mynode](times2) [right of=times, below of=times] {$\times$};
 \node[mynode](k) [left of=times, below of=times] {$k$};
 \node[mynode](m1) [left of=times2, below of=times2] {$m_1$};
 \node[mynode](m2) [right of=times2, below of=times2] {$m_2$};
 \node[mynode](r) [left of=pow, below of=pow] {$r$};
 \node[mynode](2) [right of=pow, below of=pow] {$2$};
 \draw[black] (top) to (times);
 \draw[black] (top) to (pow);
 \draw[black] (times) to (k);
 \draw[black] (times) to (times2);
 \draw[black] (times2) to (m1);
 \draw[black] (times2) to (m2);
 \draw[black] (pow) to (r);
 \draw[black] (pow) to (2);
\end{tikzpicture}\\
(b) 
\end{tabular}
\caption{\label{expressiontree}Expression trees for (a) an upper bound on square footage $x_1/x_2 + x_3$ where $x_1$ is {\it 300K}, $x_2$ is {\it pricePerSquareFoot}, and $x_3$ is {\it bathrooms} and (b) gravitational force $k m_1 m_2/r^2$. }
\end{figure}

The aim is to generate conjectured bounds that are true for any realization of input examples $E$.  The Dalmatian heuristic provides criteria for generating conjectured bounds that are the best for $E$.   Algorithm \ref{conjecturinginv} provides a way to generate expressions of increasing complexity, apply the heuristic, and store conjectures.  The {\it complexity} of an expression is the number of nodes in the corresponding expression tree (Figure \ref{expressiontree}) and is the sum of the number of invariants, number of unary operators, and number of binary operators.  The algorithm proceeds by generating unlabeled trees and then labeling the nodes with operators and invariants.  Expressions satisfying the Dalmatian heuristic conditions are retained as conjectures $\mathcal{C}$.  For a conjecture $c \in \mathcal{C}$, let $c(i)$ be the conjectured bound for example $i$.  

With examples $E$, invariants $A$, operators $O$, invariant of interest $\alpha^*$, an upper limit on the proportion of missing values allowed for an invariant $skips$, and a direction  indicating if the algorithm will produce upper or lower bounds ($UPPER$ or $LOWER$), {\bf procedure} \textsc{Conjecturing-INV} is called (Algorithm \ref{conjecturinginv}, line \ref{procedureinv}).  The number of unary nodes $u$ and binary nodes $b$ of an expression tree are initialized to zero and the conjectures $\mathcal{C}$ is initialized to the empty set  (Algorithm \ref{conjecturinginv}, line \ref{initializeinv}).  Line \ref{stopping} of Algorithm \ref{conjecturinginv} refers to the stopping criteria of the expression generator.  For invariant conjecturing for upper bounds, if the minimum conjectured bound is tight for each example (i.e., $\min_{c \in \mathcal{C}} c(i) = \alpha^*(i)$ for $i\in E$), then the expression generator is stopped.  Otherwise, expression generation continues until a time limit is reached.  {If exact bounds are not discovered for each example, more complex expressions are generated for larger time limits.}  

Line \ref{generatetree} calls a procedure to generate a tree, the branching nodes of which will be operators and the leaf nodes of which are invariants.  Lines \ref{binaryzero}-\ref{endifbinaryzero} enumerate every tree where each vertex connected to a leaf node has degree one or two.  These branching nodes will correspond to unary or binary operators, respectively, when the tree is labeled.  The leaf nodes will correspond to invariants.  Unlabeled trees are grown recursively and then the nodes are labeled with operators and invariants. 

The {\bf procedure} \textsc{generateTree} (Algorithm  \ref{conjecturinginv}, line \ref{generatetree}) creates a new tree with a single node, then calls {\bf procedure} \textsc{generateTreeRec} to add new nodes until there are $u$ unary nodes and $b$ binary nodes. The {\bf procedure} \textsc{generateTreeRec} (Algorithm \ref{conjecturinginv}, line \ref{generatetreerec}) either calls \textsc{generateLabeledTree} to apply labels by assigning invariants to leaf nodes and operators to branching nodes to generate an expression (Algorithm \ref{conjecturinginv}, line \ref{callgenlabeledtree}), or adds nodes to grow the tree (lines \ref{grow1}-\ref{grow2}). 

The {\bf procedure} \textsc{generateLabeledTree} (Algorithm \ref{conjecturinginv}, line \ref{generatelabeledtree}) takes as input a tree with $u$ unary nodes and $b$ binary nodes.  Line \ref{suffixorder} {orders the nodes so that child nodes appear before their parent}.  Then line \ref{labeltree} creates a set of labeled trees.  The leaf nodes are labeled with invariants and the branching nodes are labeled with operators.  Invariants with more than $skips$ missing values among examples are not used for labeling. For the commutative binary operators, the left child is larger than the right if the left has more nodes.  If the number of nodes is equal, we use the lexicographically largest string of labels. Since the suffix order guarantees that all subtrees are fully labeled before their parent is labeled, this is an unambiguous definition.  Examples of labeled expression trees are given in Figure \ref{expressiontree}.

Lines \ref{dalmatian} and \ref{dalmatian2} are the Dalmatian heuristic.  A conjectured upper bound $c$ is only retained in the database of conjectures $\mathcal{C}$ if the bound passes the following two tests:
\begin{enumerate}
\item (\textit{Truth test}). The candidate conjecture $\alpha^*(i) \leq c(i)$ is true for all examples $i \in E$,
and
\item (\textit{Non-dominance test.})
There is an example $i$ where $c(i) < \min\{c'(i): c' \in \mathcal{C} \setminus \{c\}\}$.
That is, the candidate conjecture would give a better bound for $\alpha^*(i)$ than any previously conjectured (upper) bound.
\end{enumerate}
Line \ref{dalmatian2} ensures that the number of conjectures is no larger than the number of examples; i.e., $|\mathcal{C}| \leq |E|$.

The procedure is the same for generating lower bounds with the only difference being how the Dalmatian heuristic criteria are evaluated in lines \ref{dalmatian} and \ref{dalmatian2}.

The computational requirements of Algorithm \ref{conjecturinginv} increase exponentially with the number of invariants and with the number of operators.  The computation time {per expression} increases with the number of examples because of the check in Step \ref{dalmatian}.  To facilitate generation of more candidate expressions in less time, one can use fewer examples as input to the algorithm.  To achieve additional efficiency, we implement the following design choice.
In our implementation, when a tree is labeled, operators can be reused, but invariants cannot.  We make this design choice so that more expressions can be generated in a smaller amount of time.  In Section \ref{nguyensec}, we will demonstrate how this limitation can be overcome in situations where repeating invariants is warranted.

\begin{algorithm}
\scriptsize
\caption{\label{conjecturinginv}Invariant Conjecturing.}
Input: Examples $E$, Invariants $A$, operators $O$, invariant of interest $\alpha^*$, invariant missing value limit $skips$, direction ($UPPER$ or $LOWER$).   \\
Output: Conjectured $\mathcal{C}$ in the form of conjectured bounds on the invariant of interest $\alpha^*$.
\begin{algorithmic}[1]
\Procedure{Conjecturing-INV}{} \label{procedureinv}
 \State {Set $u = 0$,  $b = 0$, $\mathcal{C} = \emptyset$.} \label{initializeinv}
  \While {not stopped} \label{stopping}
     \State {\textsc{generateTree}($u$, $b$).} \label{generatetree}
     \If {$b = 0$} \label{binaryzero}
         \State {Set $b = \lceil u/2 \rceil$.}
         \State {Set $u = 0$.}
     \Else
        \State {$b - -$.}
        \State {$u+=2$.}
        \EndIf \label{endifbinaryzero}
  \EndWhile
  \State \Return $\mathcal{C}$.
\EndProcedure
\Procedure {generateTree}{$u$, $b$}
   \State {Set tree = new tree with single node.}
   \State {\textsc{generateTreeRec}(tree, $u$, $b$).}
\EndProcedure
\Procedure {generateTreeRec}{tree, $u$, $b$}  \label{generatetreerec}
   \If {number of unary nodes == $u$ and number of binary nodes == $b$} 
     \State \textsc{generateLabeledTree}(tree). \label{callgenlabeledtree}
   \Else 
      \For {all nodes $v$ on the second-deepest level that have at most 1 child and have no nodes at the same level to their right with at least 1 child}  \label{grow1}
        \State  Add child to $v$.
         \State \textsc{generateTreeRec}(tree, $u$, $b$).
         \State Remove that child from $v$.
      \EndFor
      \For {all nodes $v$ on the deepest level} 
        \State Add child to $v$.
         \State \textsc{generateTreeRec}(tree, $u$, $b$).
         \State Remove that child from $v$.
\EndFor \label{grow2}
\EndIf
\EndProcedure
\Procedure {generateLabeledTree}{tree} \label{generatelabeledtree}
  \State {Order the nodes in a suffix order.} \label{suffixorder}
\State  {Recursively label each node in this ordered array with either an invariant, a unary operator, or a binary operator depending on its degree. For commutative binary operators we only label a vertex if its left child is larger than its right child.}  \label{labeltree}

   \For {each fully labeled tree} 
   \If{the corresponding bound $c$ is valid for all examples in $E$ and is not dominated by existing bounds in $\mathcal{C}$.} \label{dalmatian}
   \State{Set $\mathcal{C} = \mathcal{C} \cup {c}$.}
   \State{Remove dominated conjectures from $\mathcal{C}$.} \label{dalmatian2}
   \EndIf
   \EndFor
\EndProcedure
\end{algorithmic}

\end{algorithm}

\begin{figure}
\begin{tabular}{c}
\includegraphics[width=5.0in]{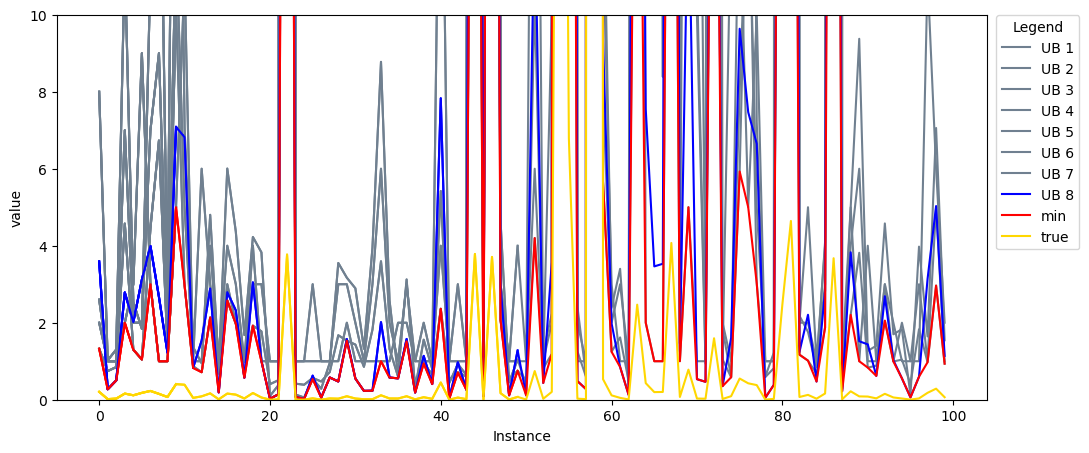} \\
(a) \\
\includegraphics[width=5.0in]{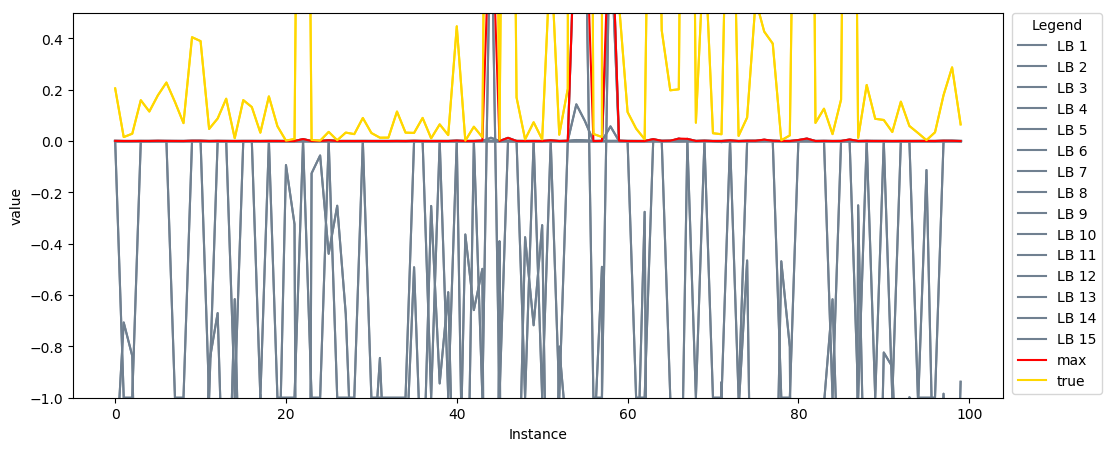}\\
(b)
\end{tabular}
\caption{\label{conjecturing_inv}(a) Upper bounds and (b) lower bounds generated for gravitational force using \textsc{Conjecturing-INV}, the invariant version of the conjecturing algorithm.  Instances from the training data are on the $x$-axis.  The gold curve is the true value for the instances. The blue curve in (a) is the true value without the constant of proportionality and is one of the upper bounds.  The red curve is the (a) maximum and (b) minimum of the discovered bounds.} 
\end{figure}

Figure \ref{conjecturing_inv} displays (a) upper bounds and (b) lower bounds derived for test instances for data generated based on a formula for gravity.  The gray curves correspond to bounds, and each must be the best on at least one training example instance in order to be retained.  More details on this experiment are provided in Section \ref{gravity} and Section \ref{regression}.

%\subsection{Property Conjecturing}
Algorithm \ref{conjecturinginv} can be adapted for property conjecturing with few modifications.  We now detail the differences.  Let $E$ be a set of examples and let $\Pi = \{\pi_1, \pi_2, \ldots, \pi_m\}$ be properties.  The examples are $n$ data observations and the properties are $m$ boolean features.  The truth value of example $i$ for property $\pi_j$ is $\pi_j(i)$.  Let $O$ be the following collection of operators: NOT ($\neg$), AND ($\&$), OR ($|$), XOR (exclusive or) ($\oplus$), and IMPLIES ($\rightarrow$).  NOT is a unary operator and the remaining operators are binary operators.  Let $\pi^* \in \Pi$ be the property for which sufficient and/or necessary conditions are of interest, and let $\pi^*(i)$ be the truth value of the property of interest for example $i$.  

The aim is to generate conjectured sufficient or necessary conditions for the property of interest that are valid for any realization of input examples $E$.  The algorithm for property conjecturing {\bf procedure} \textsc{Conjecturing-PROP} generates unlabeled trees as in Algorithm \ref{conjecturinginv} but then labels the nodes with operators and properties.  Logical expressions satisfying the Dalmatian heuristic conditions are retained as conjectures $\mathcal{C}$.  For a conjecture $c \in \mathcal{C}$, let $c(i)$ be the conjectured truth value for example $i$.  

The inputs to property conjecturing are examples $E$, properties $\Pi$, operators $O$, a property of interest $\pi^*$, and a direction ($SUFFICIENT$, $NECESSARY$) indicating if the algorithm will produce sufficient or necessary conditions for the property of interest.

The stopping criterion for property conjecturing for the case that direction is $SUFFICIENT$ is obtaining a set of conjectures where for every example with $\pi^*(i) = $ true, each example evaluates to true for at least one conjecture.  Otherwise, expression generation continues until a time limit is reached.

The Dalmatian heuristic for property conjectures is applied as follows.  A conjectured sufficient condition $c$ is only retained in the database of conjectures $\mathcal{C}$ if the expression passes the following two tests:
\begin{enumerate}
\item (\textit{Truth test}). For all examples $i \in E$ for which $c(i)$ is true, then $\pi^*(i)$ is also true, and
\item (\textit{Non-dominance test.})
The number of examples $i \in E$ for which $c(i)$ is true is not a subset of examples that evaluate to true for any previously conjectured sufficient condition.
\end{enumerate}

To generate necessary conditions for $\pi^*$, one can generate sufficient conditions for $\neg \pi^*$ (NOT $\pi^*$).

\begin{figure}
\begin{tabular}{c}
\includegraphics[width=5.0in]{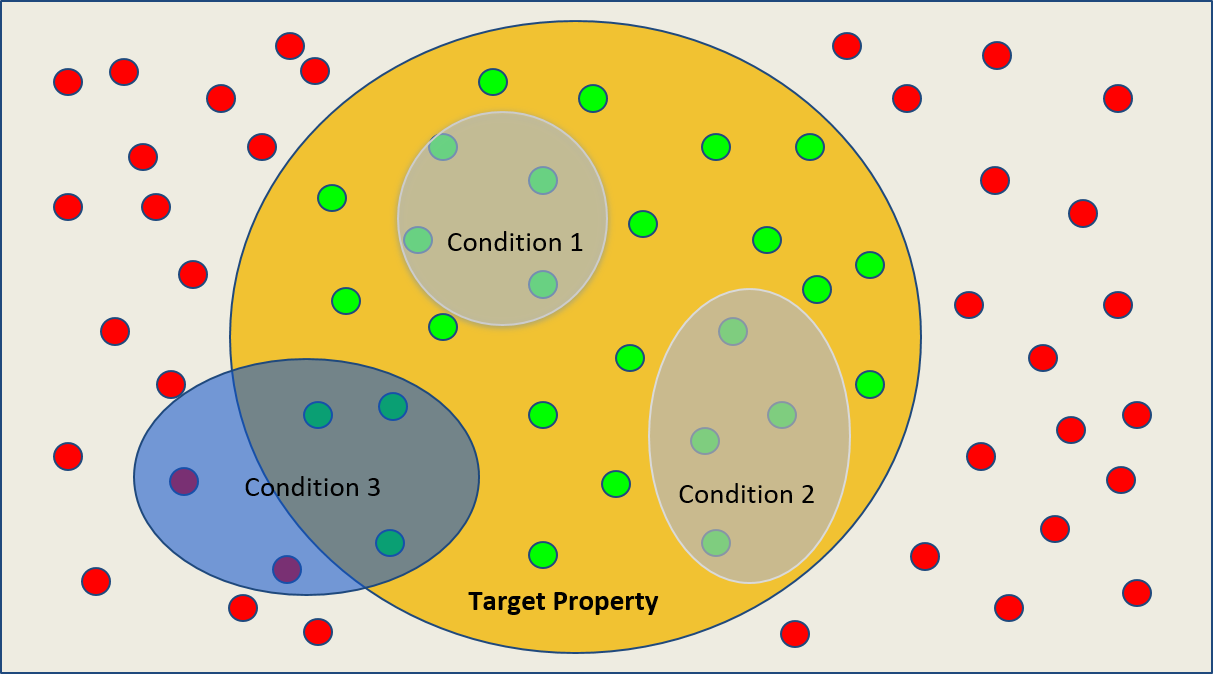} \\
(a) \\
\includegraphics[width=5.0in]{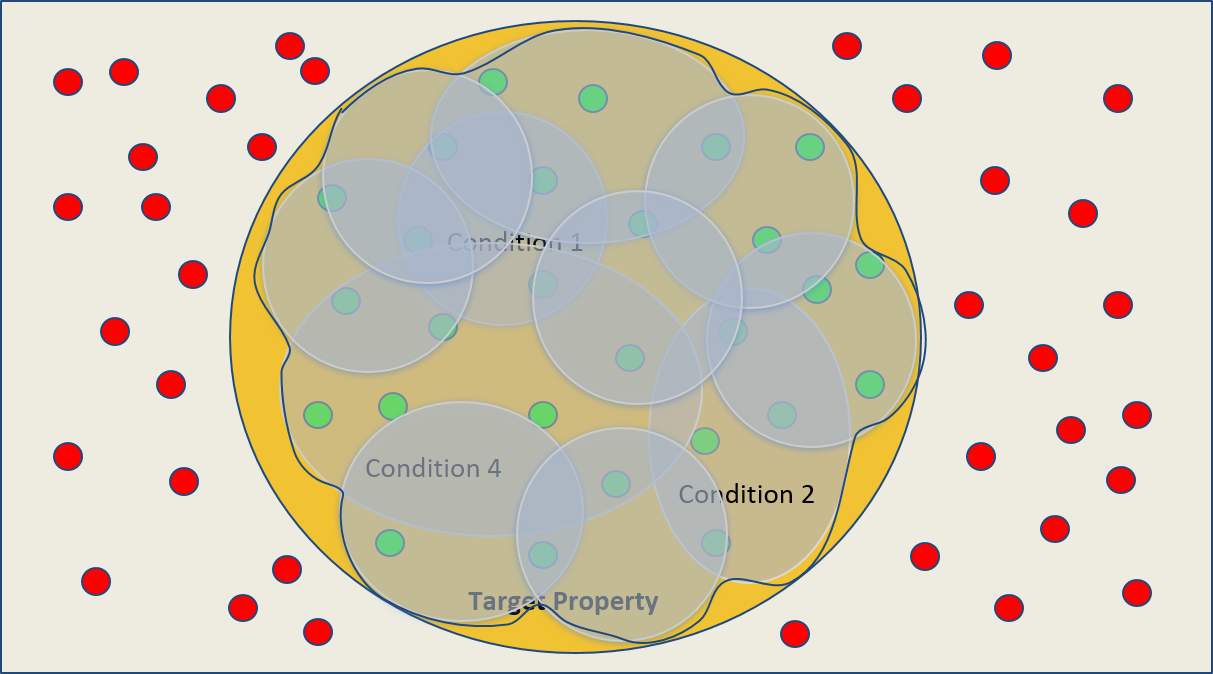}\\
(b)
\end{tabular}
\caption{\label{conjecturing_prop}Schematic of property conjecturing.  In (a), conditions 1 and 2 evaluate to true for a subset of examples with the property of interest and for no samples that do not have the property of interest.  Condition 3 evaluates to true for examples with and without the property of interest, and so it is discarded.  In (b), the union of sufficient conditions  covers all examples with the property of interest.}
\end{figure}

Figure \ref{conjecturing_prop}(a) depicts candidate conditions for examples with the property of interest (green) and those without (red).  Conditions 1 and 2 are sufficient conditions for subsets of examples with the property of interest.  Condition 3 evaluates to true for examples with and without the property of interest, and would therefore not be retained.  The goal of property conjecturing is to find a set of sufficient conditions that evaluate to true for all examples with the property of interest and for none of the examples without the property of interest as illustrated in Figure \ref{conjecturing_prop}(b).

\subsection{Other Related Work}
In this section, we explain how our work is related to previous work in automated scientific discovery, machine learning interpretability, automated feature engineering, and empirical model building.   

Symbolic regression has been used as a tool for automated scientific discovery.  Symbolic regression is the use of genetic programming to approximate a target function on training data and generalize to produce predictions on new data \citep{NicolauAgapitos20}. Until the work of \citet{SchmLips09}, the focus was on improving prediction accuracy by approximating an underlying function rather than a focus on discovering true functional relationships among features.  \cite{SchmLips09} extend previous work to develop a system for discovering laws for dynamical systems by considering relationships among derivatives.  Their work led to the development of a software \textsc{Eureqa}.  More recently, \cite{UdrescuTegmark20} combined a variety of strategies including dimensional analysis, symmetry identification, neural network training, and  brute-force enumeration into a framework called \textsc{AI Feynman} to recover true physical functional forms from data.  \citet{petersen} propose a method for deep symbolic regression that combines reinforcement learning with a recurrent neural network model.  They compare their approach with methods based on priority queue training proposed by \citet{abolafia} and the traditional genetic programming approach.  Our experiments include comparisons with each of these methods in the ability to recover equations from data.

Other frameworks have been proposed to computationally generate conjectures from data and discover scientific laws.  Data smashing is introduced by  \cite{ChattopadhyayLipson14} as a method for computing dissimilarities from streams of data (e.g., electroencephalogram data) to aid in revealing relationships among observations.  \cite{Jantzen16} proposes an algorithm with the similar purpose of detecting types of dynamical systems called {\it dynamical kinds}.  Subsequently, these kinds ``are then targets for law-like generalization'' \citep{Jantzen16}.  While Jantzen's work provides a method for discovering the kinds, it does not suggest how to recover the ``laws''.  It is these relationships that we aim to discover with the \textsc{Conjecturing} framework.  

Our work is distinguished from these previous works in that 1) we focus on generating bounds for invariants that serve as hypotheses for the investigator rather than recovering true functional forms or generating accurate predictions, 2) our invariant conjecturing algorithm is paired with a property conjecturing algorithm for discovering both nonlinear bounds and boolean relationships, 3) our framework is designed for a given static observational dataset rather than on discovering laws for dynamical systems, and 4) rather than a stochastic search over the space of functional forms, our \textsc{Conjecturing} system leverages sophisticated techniques for enumerating expressions of increasing complexity  (described in \citet{LarsVanc16} for ``noiseless'' data involving mathematical objects such as graphs).  In our system, the human remains ``in the loop'' to evaluate the plausibility of suggested bounds and conditions.

\citet{brunton} introduce \textsc{SINDy} for combining sparse regression and expert knowledge to developing models of dynamical systems.  We adopt a similar approach to incorporating prior knowledge in Section 5.4 for the Nguyen benchmark suite \citep{nguyen} where we provide the \textsc{Conjecturing} framework with candidate non-linear functions as building blocks.  Unlike our \textsc{Conjecturing} framework, theirs is designed for recovering equations governing dynamical systems rather than bounds and theirs is not capable of recovering boolean relationships.

\cite{Langley19}  provides a review of past efforts in computational scientific discovery.  Several frameworks have their origins in analyzing mass spectroscopy and other electrochemical data.  Bacon \citep{LangEtal87} is a general framework for scientific discovery based on suggesting and executing a series of designed experiments.  \cite{Tallorin18} proposed a method called POOL that uses Bayesian optimization and machine learning in an iterative fashion for experiments to discover peptide substrates for enzymes. Bacon and POOL both make recommendations regarding additional data to collect while our system assumes that a fixed dataset is provided that may or may not be the result of a designed experiment.

Precise definitions of ``explainability'' and ``interpretability'' are still being developed \citep{Vilone20, Lee20, Furnkranz20} as research in the area has rapidly accelerated.  According to the convention of \cite{Rudin19}, explainability is concerned with post-hoc analyses of black box models to create simple explanations of model behavior.  Motivated by observed accuracies of deep learning models, work in this area includes identifying important features for prediction, building simple local models, conducting sensitivity analyses, and deriving prototype examples \citep{SamekMuller19, Elton20}. 
\citet{tsang:18, tsang2:18, tsang:20} develop neural network frameworks for identifying sets of features for which there is an {\it interaction} - a non-additive relationship among predictive features that influence a response value.  These methods provide explainability in that they identify sets of features that interact, but the framework is not designed to reveal the functional form of the nonlinear interaction.

\cite{Rudin19} advocates the development of interpretable models where the mechanisms for predictions are simple relationships that are readily apparent to the investigator.  Much of the recent work in this area is in the development of decision rules (e.g., \citep{hammer:06, dash:18, gottlob:21}) or decision lists and trees (e.g., \citep{wang:15, WangRudin17, RudinErtekin18, bertsimas:17, verwer:19, blanquero:21, aghaei:21, akyuz:21}).   Different from these works, our \textsc{Conjecturing} framework automates the discovery of nonlinear features.  In addition, as with work on decision rules in general, our framework can combine the discrete features in data with the discovered nonlinear features to discover a potentially richer set of boolean relationships when compared to optimization-based trees and decision lists.

% Survey and critique of techniques for extracting rules from trained artificial neural networks \cite{AndrDiedTick95}.  cited over 1300 times.  

%F\"{u}rnkranz et al. 2020.  operational definitions of interpretability and related concepts.  experiments about plausibility, etc.

%https://www.york.ac.uk/assuring-autonomy/body-of-knowledge/implementation/2-8/interpretability-techniques/

%Samek and Muller \cite{SamekMuller19}.  Types of explanations: groups of neurons encoding a concept, explaining predictions (important features that identified an object as belonging to a class), explaining prediction strategies, representative examples.  They have developed a bunch of methods, particularly for image processing.

%\cite{dk lee} CMU Marketing/analytics guy: DK Lee (AAAI 2020)

%Article describing state-of-the-art explainability:
%https://www.zestfinance.com/snake-oil-explainability

%https://www.kdnuggets.com/2019/12/googles-new-explainable-ai-service.html
%
%https://www.kdnuggets.com/2019/12/interpretability-part-3-lime-shap.html

%Sanjeeb Dash gave a talk on MIP and interpretable models.  Paul can't find a pre-print.
%O'Reilly book on interpretability: 12/19

%Harvard biostats guy: Junwei Liu

%mueller - DARPA review

 \citet{KhuranaSamulowitzTuraga18} propose a system that leverages reinforcement learning to search expression trees for predictive features.  \textsc{ExploreKit} \citep{KatzShinSong16} is a framework for automatic feature engineering that combines features using basic arithmetic operations and then uses machine learning to rank and select those with high predictive ability.  \textsc{The Data Science Machine} \citep{KanterVeeramachaneni15} automatically generates features for entities in relational databases with possible dependencies between tables followed by singular value decomposition.  In none of these works is model transparency evaluated but rather only model performance.  An important distinction of our work from these is that they focused on improving prediction accuracy, sometimes at the expense of understandable features, and not on scientific discovery.

Traditional statistical methods for empirical model building (e.g. regression analysis) tend to focus on first- and second-order polynomial models; interaction terms up to a certain degree are often included. Empirical models are intended to provide adequate prediction performance while also providing a simple assessment of feature importance via model coefficients. Techniques such as all-subsets, stepwise selection, and regularization methods (e.g., LASSO \citep{Tibshirani96}) are commonly used to perform feature selection over model spaces of increasing complexity. However, domain knowledge is typically required for reciprocal or non-polynomial relationships. %While nonlinear regression techniques exist \citep{seber2003nonlinear, SongLangfelderHorvath13}, searching for such relationships is a departure from common practice in statistics. 
Our \textsc{Conjecturing} framework provides a search over a much broader class of nonlinear functions. 
%Google Parking app example.

\section{Two Motivating Examples}
\label{motivation}

In this section, we describe two datasets where a ``typical'' knowledge discovery workflow fails to reveal important relationships among features.  

Research on machine learning does, of course, lead to conjectured relationships between variables which are in turn used to make predictions of one or more variables in terms of others. A trained neural net, for instance, can be viewed as a black box representing a function which produces an output for every input in its domain. These functions are complex and of a different character than classical scientific laws: in particular, there is little hope of deriving these functions or relationships from simpler existing laws.  Our \textsc{Conjecturing} framework aims to help fill this gap in current capabilities. 

\subsection{Discovering Gravity}
\label{gravity}
In this example, a numeric invariant of interest is determined by a more complex nonlinear relationship with three numeric predictors.   Consider measurements including the masses of two objects $m_1$ and $m_2$, their distance $r$, and the gravitational force between them $F$.  The goal is to recover the dependence of $F$ on $m_1$, $m_2$, and $r$, or
\[
F = k \frac{m_1m_2}{r^2},
\]
where $k$ is the gravitational constant.  Following the demonstration by \citet{LangEtal87}, we create a fictional dataset using a predefined value for $k$ that is a random number between 0 and 1.  For our illustrative example, we generated 1,000 training data points and 1,000 test data points with $k=0.057098$.  Values for $m_1$, $m_2$, and $r$ are samples from Uniform(1,100000) distributions, and $F$ is calculated for each sample with no noise.

A linear regression model will fail to capture the nonlinear interaction of the variables.  Off-the-shelf machine learning methods such as random forests and neural networks can leverage the nonlinear relationship in the data but cannot present the relationship to the investigator.   In the next section, we propose a framework for producing bounds on $F$ that are functions of the other features.

\subsection{Discovering an Interaction in Real Estate Valuation Data}
\label{realestate}
The second example is a case where a boolean variable of interest is {almost} completely determined by the product of two numeric features in the dataset; i.e., the second-order \textit{interaction} term completely defines the relationship. %Although interactions of any number of variables is plausible, we only consider second-order interactions in this paper. 

Consider a dataset on residential real estate properties for sale obtained from \url{https://www.redfin.com}.  The goal is to predict whether a home with given feature values has a list price above or below \$300,000.  

%We filter data to properties with at most 20,000 square feet and priced at most \$2,000 per square foot.  The data are partitioned into equal-sized training and testing sets consisting of around 15,000 observations.  Missing data are imputed using the median from the training set.

This dataset includes both the price per square foot and total square footage along with eight additional features such as the number of bathrooms and bedrooms.  The property of interest (above vs. below) can be determined {(with some rounding error)} by multiplying the price per square foot by square footage and setting a threshold.  Thus, the interaction of price per square foot and square footage, hereafter called the {\it active interaction}, {almost} completely describes the relationship between the predictors and response.  Data are partitioned into a training dataset with 1,000 houses and a test dataset with 30,156 houses.  In the next section, we leverage our framework for invariant bounds and then extend it to produce boolean relationships to {discover} the active interaction term and how it {is related to} class membership.

{ Standard machine learning methods are able to achieve high rates of prediction accuracy and some can identify the terms of the active interaction term as important with this data, but to our knowledge, none can help the investigator discover that the terms should be multiplied.}

\section{A Conjecturing Framework for Discovering Patterns in Data}
\label{conjecturing}

We now describe a framework that leverages a conjecturing algorithm to discover nonlinear and boolean feature relationships in data.  All experiments were run on a computer with an Intel i7-2600 CPU @ 3.4GHz and 16 GB RAM.

\subsection{Conjecturing for Nonlinear Relationships}
\label{regression}
The invariant version of the conjecturing method ({\bf procedure} \textsc{Conjecturing-INV}) can be used for discovering nonlinear relationships in data.  Invariant conjectures are generated that provide upper and lower bounds on the invariant of interest.  These conjectures are the nonlinear functions that can be used as new features and/or as a complete model for the system.  

For the gravity case from Section \ref{gravity}, the invariants are $A = \{F, m_1, m_2, r\}$ and the invariant of interest is the force $F$.  The examples $E$ are the observations in the data.

%First, we mean-center the values for the response so that the intercept for any fitted model is zero (Step \ref{meancenter}).  

The \textsc{Conjecturing} framework is not designed to recover constants such as the gravitational constant $k$.  In general, for a functional relationship with a constant $k$ such that $0 < k < 1$, the expression without the constant provides {an upper bound for the response.  In cases where the constant is larger than 1, the expression without the constant provides a lower bound}. 

For our example, \textsc{Conjecturing-INV} returns 19 upper bounds and 24 lower bounds for $F$.  Among the upper bounds is 
\[
F \leq  m_1m_2/r^2,
\]
which approximates the true gravity relationship used to generate the data.  The bound does not include the constant $k$.  
%A linear regression model recovers the ``true'' gravitational constant $k=0.057098$.  
Other bounds generated by \textsc{Conjecturing-INV} include
\begin{align}
F & \leq 2m_2/\sqrt{r},\\
F & \leq 2|m_1-m_2|, \\
F & \geq 8m_2/r^2, \\
F & \geq -1/(r-2m_2).
\end{align}

Eight of the upper bounds and 15 of lower bounds for $F$ are depicted in Figure \ref{conjecturing_inv}.  The upper bound $m_1m_2/r^2$ in Figure \ref{conjecturing_inv}(a) is blue, while the true value $km_1m_2/r^2$ is gold.

As the primary goal of our approach is discovery, the bounds produced are suggestions that require further validation.  We consider it a success that the relationship $F \propto m_1m_2/r^2$ is included in one of the bounds.  Among the other bounds produced, we see that true relationships $F \propto m_1$, $F \propto m_2$, and $F \propto 1/r^2$ are all suggested, along with false relationships $F \propto 1/\sqrt{r}$ and $F \propto 1/r$.  Follow up investigations can be used to inspect these relationships and potentially recover the gravitational constant.   
An approximation of the gravitational constant of 0.057 could be represented as $(+1+1+1+1+1)\times 10^{-1-1} + (+1+1+1+1+1+1+1)\times 10^{-1-1-1}$ which by itself has complexity 22.  The expression $m_1m_2/r^2$ has complexity 6.
%+1 is not a repeated invariant, but rather a repeated operator

In this example, the \textsc{Conjecturing} framework recovers the true nonlinear relationship up to a constant of proportionality along with {42} additional suggested bounds.  Therefore, isolating a single true bound, in the case where the bound is unknown, can require additional analysis and/or experiments.  The additional bounds can provide potential insight into feature interactions.    

{To investigate the potential dependence of results on the gravitational constant $k$, we conduct an additional experiment where for each of 10 replications, a different constant $k$ is sampled from a Uniform(0,1) distribution.  For each value of $k$, \textsc{Conjecturing-INV} recovers $m_1m_2/r^2$ as a conjectured upper bound for $F$.}

{In another experiment, values for $m_1$, $m_2$, and $r$ are sampled from Uniform(1,100) distributions and the force $F$ is calculated for each sample.  Ten datasets with noise are created by sampling from a normal distribution with mean zero and standard deviation $10^{-t}$ multiplied by the root-mean-square of the $F$ values in the training data for $t=0, 1, \ldots, 9$.   Noise is added to each calculated value for $F$.  For $t\geq 4$, \textsc{Conjecturing-INV} is able to recover the true nonlinear relationship up to a constant of proportionality, among other bounds.}

\subsection{Conjecturing for Nonlinear and Boolean Relationships with Mixed Data}
\label{realestate_exp1}
Our \textsc{Conjecturing} framework for mixed data leverages the invariant version ({\bf procedure} \textsc{Conjecturing-INV}) and the property version ({\bf procedure} \textsc{Conjecturing-PROP}) of the conjecturing algorithm.  For mixed data, we propose a framework to produce conjectures of nonlinear and boolean patterns.  These conjectures can capture complex patterns while maintaining interpretability.  

We assume that we are given a dataset with numeric features $N$, boolean features $B$, and a categorical feature of interest with levels $\mathcal{Y}$.  Note that a categorical feature with more than two levels can be converted to a series of boolean features.  Let $\pi_y$ be the property that an observation has value $y$, for $y \in \mathcal{Y}$.

For each level $y\in \mathcal{Y}$, the algorithm discovers bounds for the numeric features that are satisfied by each observation in the class (Algorithm \ref{classFramework}, Lines \ref{loop1}-\ref{endloop1}).  These inequalities are converted to properties of the form ``if the inequality is satisfied, then true; false, otherwise'' (Algorithm \ref{classFramework}, Line \ref{convert}).  These new properties are combined with the original boolean features in the data (Algorithm \ref{classFramework}, Line \ref{combine}).  The properties from across all classes are pooled together and the observations belonging to all classes are pooled together as examples and then, for each level $y\in \mathcal{Y}$, the property version of conjecturing is applied to discover sufficient conditions for $\pi_y$ (Algorithm \ref{classFramework}, Lines \ref{loop2}-\ref{endloop2}).  

We now provide further details on Algorithm \ref{classFramework} using the real estate valuation case from Section \ref{realestate} as an illustrative example.  First, we convert the categorical feature {\it propertyType} into boolean features {\it condo}, {\it mobileHome}, {\it singleFamily}, {\it townhouse}, {\it multiFamily2-4Unit}, {\it multifFamily5PlusUnit}, and {\it Other}.  
We also add a feature that is a constant value of 300,000 for each observation because it is the price cutoff and call it {\it 300K}.  
The resulting $18$ features are partitioned into numeric features $N=\{${\it bedrooms}, {\it bathrooms}, {\it squareFootage}, {\it lotSize}, {\it yearBuilt}, {\it daysOnMarket}, {\it pricePerSquareFoot}, {\it hoaPerMonth}, {\it latitude}, {\it longitude}, {\it 300K}$\}$ (Algorithm \ref{classFramework}, Line \ref{invariants}) and boolean features $B = \{${\it condo}, {\it mobileHome}, {\it singleFamily}, {\it townhouse}, {\it multiFamily2-4Unit}, {\it multifFamily5PlusUnit}, {\it Other}$\}$ (Algorithm \ref{classFramework}, Line \ref{properties}).

In our training set, there are 1,000 observations that are used as examples.  For each value of the property of interest, $\{\textrm{\it below}, \textrm{\it above}\}$, the corresponding observations serve as the examples (Algorithm \ref{classFramework}, Line \ref{examples}).  For each numeric feature, upper and lower bounds on that feature are found that are functions of the other numeric features (Algorithm \ref{classFramework}, Lines \ref{upper}-\ref{lower}).  These are found by applying the invariant relations version of the conjecturing method (\textsc{Conjecturing-INV}).   For houses with property {\it below}, there are 1,280 bounds derived.  Included are plausible relations concerning house features that are seemingly irrelevant to the classification task such as 

\begin{align}
\textit{bathrooms} & \leq  2 \times \textit{bedrooms} \\
\textit {bedrooms} & \geq  \textit {bathrooms} - 1\\
\textit {lotSize} & \geq  (\textit{squareFootage} - \textit{yearBuilt})\times \textit{bedrooms}. 
\end{align}

Also included are less-interpretable bounds such as:

\begin{align}
\textit{yearBuilt} & \geq  \textit{hoaPerMonth} \times \log(10) /\log(2\times\textit{daysOnMarket})\\
\textit{daysOnMarket} & \leq  e^{e^{\sqrt{2\times \textit{lotSize}}}} \\
\textit{hoaPerMonth} & \leq  10^{2\times \textit{bathrooms}} + \textit{squareFootage}. 
\end{align}

There are also several bounds discovered that are close approximations of the relationship present in the active interaction term, including

\begin{align}
\textit{squareFootage} & \leq  \textit {300K}/\textit {pricePerSquareFoot} + \textit {bathrooms}\\
\textit{squareFootage} & \leq  \textit {300K}/\textit {pricePerSquareFoot} + \textit {bedrooms}\\
\textit{squareFootage} & \leq  \textit {300K}/\textit {pricePerSquareFoot} + \textit {daysOnMarket}\\
\textit{squareFootage} & \leq  \textit {300K}/(\textit {pricePerSquareFoot} -1) - 1\\
\textit{pricePerSquareFoot} & \leq  -\textit {300K}/(\textit {bedrooms} -\textit {squareFootage})\\
\textit{pricePerSquareFoot} & \leq  \left\lceil \textit {300K}/\textit {squareFootage} \right\rceil\\
\textit{300K} & \geq  -(\textit {bathrooms} - \textit {squareFootage}) \times \textit {pricePerSquareFoot}.
\end{align}

For houses with property {\it above}, there are 1,457 bounds derived including a mix of simple relations and less intuitive relations.  Also included are the following three relations that are nearly identical to the active interaction relation:

\begin{align}
\textit{squareFootage} & \geq  \textit{300K}/(\textit{pricePerSquareFoot}+1)\\
\textit{pricePerSquareFoot} & \geq  \textit{300K}/\textit{squareFootage} + 1\\
\textit{300K} & \leq  (\textit{pricePerSquareFoot} + 1) \times \textit{squareFootage}.
\end{align}

The resulting invariant relations are pooled together (Algorithm \ref{classFramework}, Line \ref{combineRelations}).  The invariant relations are encoded as properties (Algorithm \ref{classFramework}, Line \ref{convert}).
The original binary features from the data are also encoded as properties  for a total of $1,280 + 1,457 + 7= 2,744$ properties. Examples of encoded properties from the invariant relations are:
\begin{align}
\textit{bathrooms} & \stackrel{?}{\leq}  2 \times \textit{bedrooms} \\
(\textit{yearBuilt} & \stackrel{?}{\geq}  \textit{hoaPerMonth} \times \log(10) / \log (2\times\textit{daysOnMarket})) \\
(\textit{squareFootage} & \stackrel{?}{\leq}  \textit {300K}/\textit {pricePerSquareFoot} + \textit {bathrooms}) \\
(\textit{squareFootage} & \stackrel{?}{\geq}  \textit{300K}/(\textit{pricePerSquareFoot}+1)). 
\end{align}
These properties can be used as boolean features that indicate whether a nonlinear relationship among numeric features is satisfied for an observation.

%\[
%\begin{align}
%\textrm{propertyType} & \stackrel{?}{=} \textrm{Multi-Family} \\
%\textrm{propertyType} & \stackrel{?}{=} \textrm{SingleFamily} \\
%\textrm{propertyType} & \stackrel{?}{=} \textrm{Townhouse} \\
%\textrm{propertyType} & \stackrel{?}{=} \textrm{MobileHome} \\
%\textrm{propertyType} & \stackrel{?}{=} \textrm{Other} 
%\end{align}
%\]

The properties generated for each level $\{\textrm{\it below}, \textrm{\it above}\}$ are collected in a set $\Pi$ along with $\pi_y$ and the seven original boolean features (Algorithm  \ref{classFramework}, Line \ref{combine}). 

For each level $\{\textrm{\it below}, \textrm{\it above}\}$, apply the property version of conjecturing to the properties $\Pi$ with the training data observations serving as the examples $E$ and level as the property of interest (Algorithm \ref{classFramework}, Lines \ref{loop2}-\ref{endloop2}).  The result is a set of properties that are sufficient conditions for the levels.

\textsc{Conjecturing-PROP} returns only two properties.  They both approximate the underlying active interaction.

\begin{align}
\textit{bathrooms} & \geq -\textit{300K}/\textit{pricePerSquareFoot} + \textit{squareFootage} & \rightarrow \textit{below}\\
\textit{squareFootage} & \geq (\textit{300K}+1)/(\textit{pricePerSquareFoot}-1) & \rightarrow \textit{above}
\end{align}

An inspection of the data reveals that for some of the houses, there is some rounding error when comparing the price to the square footage multiplied by the price per square foot.  The conjecturing algorithm compensates by using invariants as error terms.  In the first property, the error term is $\textit{bathrooms}\times\textit{pricePerSquareFoot}$.  In the second property, the error term is $\textit{squareFootage} + 1$.

When these properties are applied as classification rules for predicting whether a house will be above or below \$300,000, they produce no error on the training data.  The first property misclassifies 37 of 30,156 houses in the test data for an accuracy of 0.999.  The second property misclassifies 26 houses.  The misclassified houses are due to rounding error and miscoding of data.  For example, one house in the test data is listed as having 31,248 bathrooms and another is listed as having a price of \$459.

Despite the noise and rounding error in the data, the \textsc{Conjecturing} framework {helped to discover} the active interaction term and these properties can be used as features for classifiers with near-perfect accuracy.

\begin{algorithm}
\caption{\label{classFramework}Conjecturing framework for nonlinear and boolean relationships with mixed data}

Input: Data observations $\{1,\ldots, n\}$ with numeric features $N$, boolean features $B$, and a categorical feature of interest with levels $\mathcal{Y}$; a set of invariant operators $O$ and a set of property operators $P$.\\
Output: A set of conjectured properties $\mathcal{P}$.
\begin{algorithmic}[1]
\State Set $\mathcal{P} = \emptyset$.  \hfill /* Initialize properties set. */
\State Set $A = \{\alpha_j: j \in N \}$. \hfill /* Define the set of invariants to be the original numeric features in the data. */ \label{invariants}
\State Set $\Pi = \{\pi_j: j \in B \}$. \hfill /* Define the set of properties to be the original boolean features in the data. */ \label{properties}
\For {$y \in \mathcal{Y}$} \hfill /* Loop on the levels of the categorical feature of interest */  \label{loop1}
  \State Set $\mathcal{R} = \emptyset$. \hfill /* Initialize invariant relations set. */
  \State Set $E = \{i: \pi_y\}$.  \hfill /* Define the set of observations with level $y$ as the examples. */ \label{examples}
  \For {$j \in N$} \hfill /* Loop on original numeric features. */ 
    \State Set $R_U = \mbox{\textsc{Conjecutring-INV}}(E, A, O, \alpha_{j}, UPPER)$  \hfill /* Submit examples, invariants, and the invariant of interest to the invariant version of \textsc{Conjecturing} for upper bounds. */ \label{upper}
    \State Set $R_L = \mbox{\textsc{Conjecturing-INV}}(E, A, O, \alpha_{j}, LOWER)$  \hfill /* Submit examples, invariants, and the invariant of interest to the invariant version of \textsc{Conjecturing} for lower bounds. */ \label{lower}
    \State Set $\mathcal{R} = \mathcal{R} \cup R_U \cup R_L$. \label{combineRelations}
  \EndFor 
  \State Convert the new invariant relations $\mathcal{R}$ into properties $\Pi_\mathcal{R}$. \label{convert}
  \State Set $\Pi = \Pi \cup \pi_y \cup \Pi_\mathcal{R}$. \hfill  /* Define the set of properties to be the original boolean features, the level $y$, and the invariant relations properties. */  \label{combine}
  \EndFor \label{endloop1}
\For {$y \in \mathcal{Y}$} \hfill /* Loop again on the levels of the categorical feature of interest. */  \label{loop2}
  \State Set $E = \{1,\ldots, n\}$. \hfill /* Use all examples. */
  \State Set $\mathcal{P}_S = \mbox{\textsc{Conjecturing-PROP}}(E, \Pi, P, \pi_y, SUFFICIENT)$.  /* Submit examples, properties, and the level $y$ as the property of interest to the property version of \textsc{Conjecturing} for sufficient conditions. */ \label{property}
  \State Set $\mathcal{P}_N = \mbox{\textsc{Conjecturing-PROP}}(E, \Pi, P, \pi_y, NECESSARY)$. /* Submit examples, properties, and the level $y$ as the property of interest to the property version of \textsc{Conjecturing} for necessary conditions. */ \label{property2}
  \State Set $\mathcal{P} = \mathcal{P} \cup \mathcal{P}_S \cup \mathcal{P}_N$.
\EndFor \label{endloop2}
\State \Return $\mathcal{P}$.
\end{algorithmic}
\end{algorithm}

Hereafter, we use ``\textsc{Conjecturing} framework'' to imply:
\begin{enumerate}
    \item In the case that all features are numeric, apply {\bf procedure} \textsc{Conjecturing-INV}.
    \item In the case that all features are categorical, convert the features to a series of properties (boolean features) and apply {\bf procedure} \textsc{Conjecturing-PROP}.
    \item In the case of mixed data, apply Algorithm \ref{classFramework}.
\end{enumerate}
If there is no invariant of interest or no property of interest, each invariant and/or property can serve as the invariant/property of interest in turn, and conjectures can be generated for each.

\section{Additional Computational Experiments}
\label{complexity}

\subsection{Sensitivity to the Number of Features}

To investigate the impact of the number of features on the performance on the \textsc{Conjecturing} framework, we conduct experiments adding noisy features to the gravity example.  We use the same experimental setup as described in Section \ref{regression} including a time limit of five seconds.  For ${t}=0,\ldots,10$, we add ${t}$ noise invariants generated from a standard normal distribution and check 
\begin{enumerate}
    \item whether \textsc{Conjecturing-INV} recovers $m_1m_2/r^2$, 
    \item the number of conjectures produced, 
    \item the number of expressions evaluated, and 
    \item the number of valid expressions produced.  Valid expressions are bounds that are valid for all training examples.  
\end{enumerate}

For ${t}=0, \ldots, 6$, \textsc{Conjecturing-INV} recovers $m_1m_2/r^2$ as a conjectured upper bound and for ${t}=7, \ldots, 10$ the bound is not recovered {in the five second time limit} because of the additional noise invariants.  The number of original invariants is five including the force $F$, so in this experiment we can add more than 100\% additional invariants and still recover the true proportionality relationship $F \propto m_1m_2/r^2$.  

\begin{figure}
\includegraphics[width=5.0in]{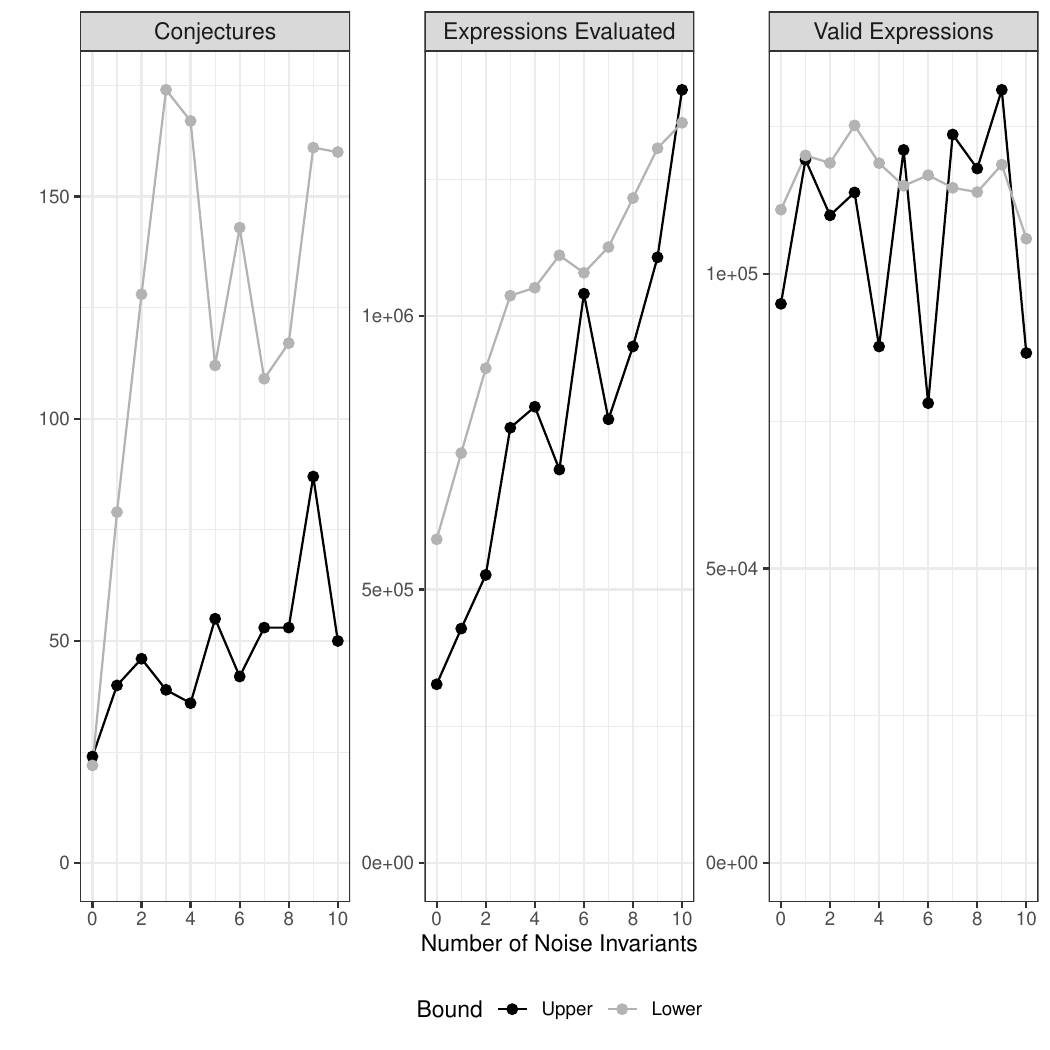} 
\caption{\label{noisetest}The number of conjectures produced, expressions tested, and valid experessions produced as a function of the number of noise invariants added to the gravity experiment.}
\end{figure}

Figure \ref{noisetest} contains plots of the number of conjectures produced, number of expressions evaluated, and the number of valid expressions produced within the five second time limit.  The number of conjectures produced and expressions evaluated increases as the number of columns increases and tends to be larger for lower bounds than upper bounds.  The number of conjectures { produced} ranges from 22 to 174.  The number of valid expressions fluctuates between 75,000 and 135,000 and there is no discernible pattern effect of the number of noise invariants ${t}$.   As ${t}$ increases, there are more invariants available and the number of low-complexity expressions increases exponentially as does the number of low-complexity expressions comprised of the noise invariants.  Low-complexity expressions can be generated and checked more quickly which is why the number of expressions evaluated increases with ${t}$.  The number of expressions evaluated in five seconds ranges between about 300,000 and 1.4 million.

\subsection{Sensitivity to Training Examples}

To investigate the effect of different subsets of training examples on the ability of the \textsc{Conjecturing} framework to recover true relationships, we apply the framework to the real estate experiment with 10 random samples of 1,000 training examples.   We use the same experimental setup as described in Section \ref{classFramework} including a time limit of five seconds.

Table \ref{trainsens} contains the conjectures produced by the \textsc{Conjecturing} framework for each of the 10 replications.  As in the experiment in Section \ref{realestate_exp1}, the \textsc{Conjecturing} framework makes use of invariants and operators to compensate for rounding error.  The invariants employed as tolerances are {\it bathrooms} and {\it longitude}.  The framework also employs operators $+1$, $-1$, and $\lceil \cdot \rceil$ to account for deviations from the underlying active interaction.  Each of the bounds can be rewritten in terms of {\it squareFootage} $\times$ {\it pricePerSquareFoot} $-300K$ plus or minus a small error term containing at most one additional invariant.  Therefore, we see that in this instance, the method is not sensitive to the choice of training examples.  Further experiments are needed to understand how well the \textsc{Conjecturing} framework can recover underlying relationships in the presence of different kinds of noise.  This is the subject of future work.

\begin{table}
\caption{\label{trainsens}Sufficient Conditions Produced by {\textsc{Conjecturing}} for Different Training Set Samples for the Real Estate Valuation Data}
\begin{tabular}{c|l}				
Sample	&	Conjectures Produced	\\	\hline \hline
1	&	$bathrooms \geq -300K/pricePerSquareFoot + squareFootage \rightarrow below$	\\	
	&	$squareFootage \geq (300K+1)/(pricePerSquareFoot-1)) \rightarrow above$	\\	\hline
2	&	$squareFootage \leq \lceil300K/pricePerSquareFoot\rceil \rightarrow below$	\\	
	&	$squareFootage \geq (300K+1)/(pricePerSquareFoot-1) \rightarrow above$	\\	\hline
3	&	$bathrooms \geq -300K/pricePerSquareFoot + squareFootage \rightarrow below$	\\	
	&	$squareFootage \geq (300K+1)/(pricePerSquareFoot-1)) \rightarrow above$	\\	\hline
4	&	$bathrooms \geq -300K/pricePerSquareFoot + squareFootage \rightarrow below$	\\	
	&	$squareFootage \geq (300K+1)/(pricePerSquareFoot-1)) \rightarrow above$	\\	\hline
5	&	$bathrooms \geq -300K/pricePerSquareFoot + squareFootage \rightarrow below$	\\	
	&	$squareFootage \geq (300K+1)/(pricePerSquareFoot-1)) \rightarrow above$	\\	\hline
6	&	$squareFootage \leq \lceil 300K/pricePerSquareFoot \rceil \rightarrow below$	\\	
	&	$squareFootage \geq (300K+1)/(pricePerSquareFoot-1) \rightarrow above$	\\	\hline
7	&	$squareFootage \leq (300K+longitude)/pricePerSquareFoot \rightarrow below$	\\	
	&	$squareFootage \geq (300K+1)/(pricePerSquareFoot-1)) \rightarrow above$	\\	\hline
8	&	$bathrooms \geq -300K/pricePerSquareFoot + squareFootage \rightarrow below$	\\	
	&	$squareFootage \geq (300K+1)/(pricePerSquareFoot-1)) \rightarrow above$	\\	\hline
9	&	$squareFootage \leq (300K-daysOnMarket)/pricePerSquareFoot \rightarrow below$	\\	
	&	$squareFootage \geq (300K+1)/(pricePerSquareFoot-1) \rightarrow above$	\\	\hline
10	&	$bathrooms \geq -300K/pricePerSquareFoot + squareFootage \rightarrow below$	\\	
	&	$squareFootage \geq (300K+1)/(pricePerSquareFoot-1)) \rightarrow above$	\\	\hline
\end{tabular}				

\end{table}

\subsection{Comparison to \textsc{AI Feynman} \citep{UdrescuTegmark20}}
\label{comparison}
In this section, we compare the ability of \textsc{Conjecturing} to recover equations from datasets used by \cite{UdrescuTegmark20} with their algorithm \textsc{AI Feynman}. 
We then apply the implementation of \textsc{AI Feynman} to the gravity and real estate datasets described in Section \ref{motivation}.  We note that the primary goal of \textsc{Conjecturing} is for discovery of nonlinear and boolean relationships while the primary goal of \textsc{AI Feynman} is recovery of equations.

{\bf Performance on Feynman Equations.} % in caption of Table 1 Udrescu and Tegmark mention using three subsets of operators
We apply \textsc{Conjecturing} to the first 10 equations listed in Table 4 of \citep{UdrescuTegmark20} to draw comparisons based on solution time and noise tolerance.  We used the data published by the authors here: \url{https://space.mit.edu/home/tegmark/aifeynman.html}.
As in \citep{UdrescuTegmark20}, for each instance we apply \textsc{Conjecturing} with three subsets of operators in turn: $\{+, -, \times, \div, +1, -1, ^2, \sqrt{}\}$, 
$\{+, -, \times, \div, +1, -1, ^2, \sqrt{}, \sin, \cos, \ln, ^{-1}, e^{}\}$, 
$\{+, -, \times, \div, +1, -1, ^2, \sqrt{}, \sin, \cos, \ln, ^{-1}, e^{}, |\cdot|, \sin^{-1}, \tan^{-1}\}$. 
For instances where an equation includes the constant $\pi$, we include $\pi$ as a constant invariant.  For each instance, we use the first 10 samples in each dataset and run \textsc{Conjecturing} for {7,200} seconds { which is the same time limit used for \textsc{AI Feynman} and \textsc{Eureqa} reported in \citep{UdrescuTegmark20}}.  We also run \textsc{Conjecturing} for the noise tolerance of and time required by \textsc{AI Feynman} to recover the equations as reported in Table 4 of \citep{UdrescuTegmark20}.  

Tables \ref{aif} and \ref{aif2} contain the results of applying \textsc{Conjecturing} to the datasets.   \textsc{Conjecturing} produces bounds that match the equation for five of the ten instances.  \textsc{Conjecturing} finds a match for all equations with complexity 10 or less, {matches one equation with complexity 11,}  and is unable to find a match for equations with higher complexity.  \citet{UdrescuTegmark20} report that \textsc{AI Feynman} resolves all of the equations while \textsc{Eureqa} \citep{SchmLips09} resolves four of the ten equations, two of which are different from those found by \textsc{Conjecturing}.  These results indicate that \textsc{Conjecturing} is well suited for recovering equations of complexity 10 or less within 7,200 seconds.  Higher-complexity formulas with more invariants require additional time.

Equations I.6.2 and I.6.2.b in Table \ref{aif} each have a repeated invariant $\sigma$.  As noted in Section \ref{dalmatiansec}, \textsc{Conjecturing} does not allow repeated invariants and so these equations will not be recoverable as bounds.  In Section \ref{nguyensec}, we describe ways to address this deficiency and recover equations such as I.6.2 and I.6.2.b.

The normalized root-mean-square error (NRMSE) calculated for 100 test examples for the best-performing conjecture on the training data based on mean absolute error. 
NRMSE is calculated as $1/\sigma_f$ multiplied by the root-mean-square error on the test examples, where $\sigma_f$ is the standard deviation of the invariant of interest for the test examples.  \textsc{Conjecturing} produced conjectures with NRMSE less than 1.0 for 8 of the 10 equations. {\citet{UdrescuTegmark20} report exact recovery of all of the equations by \textsc{AI Feynman}, so the NRMSE is 0.}
%Instance I.6.2.a has complexity 9, comprised of 2 invariants, 1 binary operator, and 6 unary operators.  

Despite the fact that \textsc{Conjecturing} is not designed for recovery of equations, we see that it can be successful in doing so for lower-complexity nonlinear equations.
%Instance I.10.7 has complexity 9, comprised of 3 invariants, 3 binary operator, and 3 unary operators.  

Table \ref{aif2} in the Appendix contains the results of applying \textsc{Conjecturing} to the datasets when noise is added to the invariant of interest.  The noise { added to the invariant of interest is sampled from a normal distribution with mean zero and standard deviation equal to the number in the table multiplied by the root-mean-square of the invariant of interest in the training data.}  The noise is the noise level tolerated by \textsc{AI Feynman} as reported by \citet{UdrescuTegmark20}.  The time is the time reported by \citet{UdrescuTegmark20} for \textsc{AI Feynman} to recover the equation.

\textsc{Conjecturing} is unable to achieve exact recovery of the equations with the introduction of noise.  The NRMSE is less than 1.0 for six of the 10 equations and does not exceed 1.625.  { Noise that results in target values above or below the ground truth will violate exact upper or lower bounds that could be produced by the Dalmatian heuristic.  Even so,} \textsc{Conjecturing} is able to produce good approximations of the equations.

{\bf Performance of AI Feynman \citep{UdrescuTegmark20} on Gravity and Real Estate Examples.}  We apply the implementation of \textsc{AI Feynman} available here: \url{https://github.com/SJ001/AI-Feynman} to the gravity example described in Section \ref{gravity} and the real estate example described in Section \ref{realestate}.  A difference between our gravity example and the datasets used by \cite{UdrescuTegmark20} is that for the gravity example, the gravitational constant is the same for every data point, but for the datasets used by \cite{UdrescuTegmark20}, it is treated as a variable and is different for each point.   

For the real estate data, we apply \textsc{AI Feynman} to the data to attempt to discover the relationship between the property of interest and the input features.  The original numeric features are supplied along with boolean features corresponding to the levels of the \textit{propertyType} feature.  Note that \textsc{AI Feynman} is designed for recovering numeric functions and is therefore not suitable for boolean relationships such as those in the real estate example.  

For both instances, the \textsc{AI Feynman} implementation aborts with an error regarding an eigenvalue calculation.  We suspect that the source of the failure in both cases may be due to the difference in treatment of constants.  In our gravity example, the gravitational constant is the same for all points while in analogous examples, \cite{UdrescuTegmark20} treat constants as variables and generate a unique value for each observation.  Our practice of treating the gravitational constant as the same for all observations may be contributing to an error in matrix calculations for \textsc{AI Feynman}.  In the real estate example, each observation has a feature with the same value (the \$300,000 cutoff).  This constant column in the data matrix could also be contributing to an error in matrix calculations for \textsc{AI Feynman}.  In the electronic companion, we include the code and output for \textsc{AI Feynman} applied to 1) their Example 1, demonstrating that our installation is functional, 2) our gravity example, including the error message, and 3) our real estate example, including the error message.  These examples show that the \textsc{Conjecturing} framework can provide useful insights on examples where \textsc{AI Feynman} cannot.

\begin{table}
{\footnotesize
\caption{\label{aif}Results for \textsc{Conjecturing} on Datasets from \citep{UdrescuTegmark20}}
\begin{tabular}{lllllll}
         &          &  Number of  &                      &       & Recovered\\
Instance & Equation &  Invariants & Complexity  & NRMSE & by \textsc{Eureqa}? \\
\hline\hline
I.6.2.a & $f=e^{-\theta^2/2}/\sqrt{2\pi}$                           & 2 & 9   & 0.000 & No\\
I.6.2 & $f=e^{-\theta^2/2\sigma^2}/\sqrt{2\pi\sigma^2}$             & 3 & 13  & 0.553 No\\
I.6.2.b & $f=e^{-(\theta-\theta_1)^2/2\sigma^2}/\sqrt{2\pi\sigma^2}$& 5 & 16  & 1.511 & No\\
I.8.14 & $d=\sqrt{(x_2-x_1)^2+(y_2-y_1)^2}$                         & 4 & 10  & 0.000 & No\\
I.9.18 & $\frac{Gm_1m_2}{(x_2-x_1)^2+(y_2-y_1)^2+(z_2-z_1)^2}$      & 9 & 17  & 1.211 & No\\
I.10.7 & $m=\frac{m_0}{\sqrt{1-\frac{v^2}{c^2}}}$                   & 3 & 9   & 0.000 & No \\
I.11.19 & $A=x_1y_1+x_2y_2+x_3y_3$                                  & 6 & 11  & 0.696 & Yes\\
I.12.1 & $F=\mu N_n$                                                & 2 & 3   & 0.000 & Yes\\
I.12.2 & $F=\frac{q_1q_2}{4\pi\epsilon r^2}$                        & 5 & 12  & 0.671 & Yes\\
I.12.4  & $E_f = \frac{q_1}{4\pi\epsilon r^2}$                      & 4 & 11  & 0.000 & Yes \\ \hline
\end{tabular}}

\end{table}

\subsection{Experiments with the Nguyen Benchmark Suite \cite{nguyen}}
\label{nguyensec}
We apply our invariant conjecturing implementation to the Nguyen benchmark suite \citep{nguyen} so as to draw comparisons with symbolic regression methods described by \cite{petersen}.  The Nguyen benchmark suite is a set of 12 equations.  As mentioned before, our \textsc{Conjecturing} framework is designed for discovering nonlinear relationships in the form of bounds and boolean relationships while symbolic regression methods are designed to recover equations.  In these experiments, we investigate the ability of invariant conjecturing to recover equations (or approximations) among the discovered bounds.  

The benchmark equations are in Table \ref{nguyentable}.  We generate an instance for each using the protocols described by \cite{petersen}.  For each equation, 20 training examples and 20 test examples are generated.  For equations 1 though 6, $x$ is sampled from a Uniform(-1,1) distribution; for equation 7, $x$ is sampled from a Uniform(0,2) distribution; for equation 8, $x$ is sampled from a Uniform(0,4) distribution; for equations 9 through 12, $x$ and $y$ are sampled from a Uniform(0,1) distribution.  For each equation, we allow a time limit of 10,000 seconds for generating upper and lower bounds.  The operators include unary operators sine, cosine, natural log, and natural exponential, and binary operators addition, subtraction, multiplication, and division. 

As noted in Section \ref{dalmatiansec}, our \textsc{Conjecturing} framework does not allow repeated invariants in conjectures.  Therefore, most of the equations in Table \ref{nguyentable} cannot be recovered by our framework.  For each equation, we first evaluate the ability of the conjectured bounds to approximate the equation by reporting the normalized root mean squared error (NRMSE) as defined and reported by \citet{petersen} and described in Section \ref{comparison}.  We report NRMSE for the conjecture with the lowest mean average error for the training examples.  

The results in Table \ref{nguyentable} indicate that our \textsc{Conjecturing} framework is able to recover only equation 11 $f=x^y$.  It is able to do so despite the fact that the exponent operator is not included.  \textsc{Conjecturing} produces the expression $e^{y\log(x)}$ and simplified it to $x^y$.  The NRMSE values for the best bounds for the other instances range from 0.22 to 1.08.  These values are larger than those reported for symbolic regression methods as reported in Table 10 of \cite{petersen}; the methods include deep symbolic regression (DSR) \citep{petersen}, priority queue training (PQT) \citep{abolafia}, vanilla policy gradient (VPG) \citep{petersen}, genetic programming (GP), and a method implemented in Mathematica based on Markov chain Monte Carlo and nonlinear regression.

For expressions 1 through 8, the only invariants are the invariant of interest $f$ and the input invariant $x$.  Because of the fact that no invariants can be repeated, the \textsc{Conjecturing} framework is limited to the application of only unary operators to $x$ and no expressions with binary operators are produced.  As an example, for the first equation, the best conjectured lower bound is
\[
f(x) \geq \sin(e^{e^{\cos(\sin(\sin(\sin(\sin(\log(\sin(\sin(\cos(\cos(\sin(e^x)))))))))))}}).
\]

We now describe how our \textsc{Conjecturing} framework can be adapted to allow for repeated invariants, and report results for the adapted method.  To address repeated invariants, we can add additional invariants using commonly-occurring functional forms.  For equations 1 through 8, we add invariants $x^2$, $x^3$, $x^4$, $x^5$, $x^6$, $\sin(x)$, $\cos(x)$, $\sqrt{x}$ along with two copies of the constant 1, and the constant 2.  Two copies of the constant 1 are included because it appears twice in equation 7.  For equations 9 through 12 we also add $y^2$ as an invariant.  Recall that in our implementation, while invariants cannot repeat in a conjectures, operators can.  Therefore, an alternative approach to addressing the constants is to include the unary operator of addition by 1.  

As shown in Table \ref{nguyentable}, the \textsc{Conjecturing} framework is able to exactly recover each equation as a bound when the additional invariants are included so that the NRMSE values are 0.000 for all equations.

The practice of adding the invariants that are nonlinear functions of the original input might appear to be impractical.  However, as suggested by \citet{brunton}, specifying these invariants can reflect expert knowledge on the system being investigated.  They note that identifying candidate functions for SINDy ``must be a coordinated effort to incorporate expert knowledge, feature extraction, and other advanced methods.''   \textsc{Conjecturing} offers distinct capabilities for discovery, as nonlinear functions can be specified as invariants or may  still be discovered so long as they do not involve repeated input invariants.

\begin{table}
\caption{\label{nguyentable}Results for \textsc{Conjecturing} on the Nguyen Benchmark Suite \cite{nguyen}}
    \centering
    \begin{tabular}{rccc|cc}
            &            & \multicolumn{2}{c|}{Without Additional Invariants} & \multicolumn{2}{c}{With Additional Invariants} \\
            \hline
   Instance &  Equation  & Recovered? & NRMSE                                &  Recovered?  & NRMSE                            \\
    \hline\hline
   1 & $f=x^3+x^2+x$                    & No & 0.94  & Yes & 0.00  \\
   2 & $f=x^4+ x^3+x^2+x$               & No & 0.86  & Yes & 0.00 \\
   3 & $f=x^5 + x^4+ x^3+x^2+x$         & No & 1.01  & Yes & 0.00 \\
   4 & $f=x^6 + x^5 + x^4+ x^3+x^2+x$   & No & 0.63  & Yes & 0.00 \\
   5 & $f=\sin{(x^2)}\cos{(x)} - 1$         & No & 0.22  & Yes & 0.00  \\
   6 & $f=\sin{(x)} + \sin{(x+x^2)}$        & No & 0.44  & Yes & 0.00 \\
   7 & $f=\log{(x+1)} + \log{(x^2 + 1)}$    & No & 0.58  & Yes  & 0.00 \\
   8 & $f=\sqrt{x}$                     & No & 1.08  & Yes & 0.00 \\
   9 & $f=\sin{(x)} + \sin{(y^2)}$          & No & 1.03  & Yes  & 0.00 \\
   10 & $f=2\sin{(x)}\cos{(y)}$             & No & 0.60  & Yes & 0.00  \\
   11 & $f=x^y$                         & Yes & 0.00 & Yes & 0.00 \\
   12 & $f=x^4-x^3+\frac{1}{2} y^2 - y$ & No & 0.83  & Yes  & 0.00 \\ \hline
    \end{tabular}
\end{table}

\section{Application to COVID-19 Data}
\label{covid}
In this section, we demonstrate the \textsc{Conjecturing} framework on synthetic patient-level COVID-19 data that was provided as part of the Veterans Health Administration (VHA) Innovation Ecosystem and precisionFDA COVID-19 Risk Factor Modeling Challenge (\url{https://precision.fda.gov/challenges/11/view}).  The data include synthetic veteran patient health records including medical encounters, conditions, medications, and procedures.   All subjects are located in Massachusetts.  

The goal of the challenge was to better understand risk and protective factors for COVID-19 outcomes.   Participants were asked to predict alive/deceased status.  { In our experiments, we focused on investigating outcomes for subjects who had COVID-19.}  Since our goal is to discover potential risk and protective factors, we evaluate the performance of \textsc{Conjecturing} by checking the performance of the feature relationships on holdout test data rather than on prediction accuracy.  Establishing the risk and protective factors as causal would require additional controlled experiments.

Predictions were based on information obtained through December 31, 2019.  In the training data, we drop all information pertaining to events on or after January 1, 2020 and drop subjects who died before January 1, 2020.  {The prediction horizon is January 1, 2020 through May 31, 2020, and there are 5,568 patients with deceased status.}   

{Table \ref{covidfeatures} in the Appendix includes definitions of new features that we generated for each patient.}  For each numeric observation, we created invariants for the mean and most recent value.  For each reported allergy, device, immunization, procedure, and discretely-measured observation we create a property corresponding to each level.  In total, we use 309 invariants and 362 properties.  We use a training set consisting of 100 subjects from each outcome class (deceased/alive).  We compare the results of applying \textsc{Conjecturing} with classification and regression trees (CART) \citep{breiman84} which is another interpretable method.  

\subsection{Results for \textsc{Conjecturing}}

Upper and lower bounds are generated for each invariant, and for each outcome.   These bounds, along with the 362 properties in the data, are used as properties for \textsc{Conjecturing-PROP}.  Conjectures are generated for both outcomes.  The parameter $skips$ is set to 90\%.  We use the remaining 73,497 subjects as a test data set thereby allowing us to ascertain the effects of potential overfitting to the 200 subjects used for training.  

Among those with COVID-19 in the test data, 5,468 (8.0\%) have a status of deceased, and 68,029 (92.0\%) are alive.  There are 38 conjectures for sufficient conditions for alive status and 40 conjectures for sufficient conditions for deceased status produced by the framework.  Tables \ref{alive}-\ref{deceasedstats} in the Appendix contain the conjectures and evaluations.

Tables \ref{alivestats} and \ref{deceasedstats} in the Appendix contain quantitative evaluations of the performance of the conjectures.  Each table contains the precision, support, and lift of each conjecture.  Note that each conjecture is a sufficient condition expressed as a conditional statement.  The precision is the percentage of test examples for which the conditional statement evaluates to true among those for which the antecedent is true.  Precision may be thought of as the ``hit rate'' of the conjecture.  The support is the number of test examples for which the antecedent evaluates to true.  The lift is the ratio of the precision to the proportion of examples for which the consequent is true.  If the lift is greater than 1, then the conjecture is better at identifying people for which the consequent is true than a random selection from the population.  

Of the 38 conjectures for alive status, 22 (57.9\%) have lift at least 1.00.  The lift ranges from 0.81 to 1.07; note that the maximum possible lift for a conjecture for alive status is 1.08 (1/(68029/73497)).  Of the 40 conjectures for deceased status, 34 (85\%) have lift at least 1.00.  The lift ranges from 0.17 to 4.15.     

Consider the sufficient conditions for deceased status in Table \ref{deceased} in the Appendix.  The conjecture with the highest precision and lift is
\[
\textit{longitude} > -\textit{age} \times \textit{medicationsLifetimePercCovered}  \rightarrow \textit{Deceased},
\]
and has a lift of 4.15 meaning that a subject for which  $\textit{longitude} > -\textit{age} \times \textit{medicationsLifetimePercCovered}$ is 4.15 times as likely to die as a randomly selected subject.  The conjecture indicates that subjects in the east who are older and have a larger percentage of medications covered by the payer are at higher risk of death.   
The presence of {\it longitude} in the conjecture could be an indication of higher risk in population centers in the east such as Boston { or it could just serve the purpose of a number so that the relationship of the other invariants is satisfied.  The range for {\it longitude} is $(-73.49,-69.92)$.  We conduct a follow up $t$-test for a difference in mean {\it longitude} by outcome and the null hypothesis is not rejected ($p=0.42$), suggesting that {\it longitude} is serving as a tolerance factor.}
The percentage of medications covered by the VHA is higher for subjects with more preexisting conditions and for those with more expensive medications because there is a low copay annual cap (currently \$700\footnote{\url{ https://www.va.gov/health-care/copay-rates/}, accessed July 10th, 2022}).
Further, the conjecture produces a suggestion of a functional form for the relationship between these factors.  
The conjecture confirms the CDC guidance that older subjects and those with more preexisting conditions are a higher risk of death from Covid\footnote{\url{https://www.cdc.gov/coronavirus/2019-ncov/need-extra-precautions/people-with-medical-conditions.html}, accessed July 10th, 2022}.  

The conjecture with the second-highest precision and lift is
\[
\textit{medicationsActive} > \lfloor \textit{hemoglobinA1cHemoglobinTotalInBlood} \rfloor \rightarrow \textit{Deceased}, 
\]
with a lift of 3.29.  The condition includes the number of active medications and the ratio of hemoglobin A1c to total hemoglobin (an HbA1c test).   The conjecture indicates that those with more active medications than the HbA1c percentage are at higher risk of death, which again agrees with the CDC guidance concerning preexisting conditions.  Typical values for HbA1c for non-diabetic patients are below 5.7\%, while diabetic subjects can have values between 6.5\% and 10.0\%    \footnote{https://www.cdc.gov/diabetes/managing/managing-blood-sugar/a1c.html, accessed July 10th, 2022}.  The number of active medications can indicate a larger number of preexisting conditions, and the conjecture suggests that for diabetic patients, those with additional conditions are at higher risk.   

The conjecture with the third-highest precision and lift is 
\[
\textit{age} > \textit{carbonDioxide}\times \lfloor \textit{potassium} \rfloor \rightarrow \textit{Deceased},
\]
with a lift of 3.12.  The conjecture suggests that older subjects with lower $CO_2$ levels and lower potassium are at higher risk of death.  Lower $CO_2$ levels and abnormal potassium levels, particularly lower levels, has been independently studied and associated with COVID-19 morbidity and mortality \citep{hu:21,noori:22}.  In addition to validating a role for these invariants, the conjecture suggests a potential nonlinear relationship among them.

For both outcomes, the \textsc{Conjecturing} framework is able to generate new sufficient conditions that are true for the respective outcome at higher rates than would be expected for a patient selected at random.  These results indicate that the conjecturing process is capturing relationships that hold across the population and are not merely reflective of the 200 training samples.  In other words, overtraining appears to be mitigated.  
The discovered relationships, and the direct and indirect relationships that they indicate among features, are validated by the medical literature and provide suggestions for deeper investigations into the functional form of the relationships and the extent of causality.  
The number of conjectures generated is not overwhelming for a human investigator to consider and further investigate.  

\subsection{Comparison with an Interpretable Model}
\begin{figure}
\begin{center}
\includegraphics[width=5.0in]{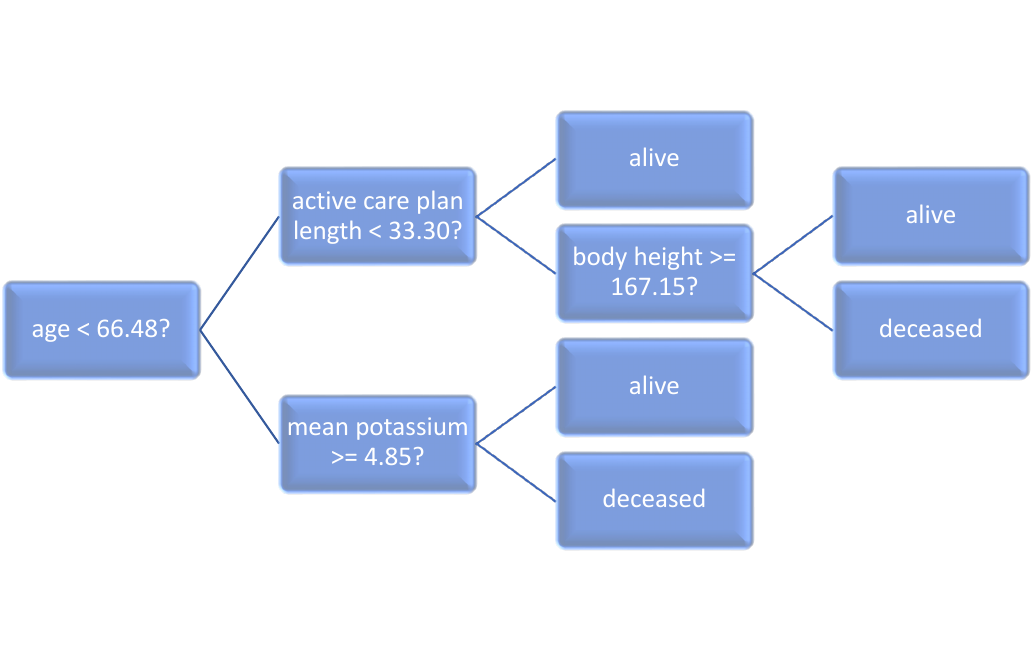} 
\end{center}
\caption{\label{tree} {Tree produced by CART for predicting alive/deceased status for COVID-19 patients.  For each node, if the condition is satisfied, then the upper branch is taken.}} 
\end{figure}

We now consider the results of applying classification and regression trees (CART)  \citep{breiman84} to the COVID-19 data.  A model is fit using the implementation in the R library {\it rpart} \citep{rpart}.  The tree produced by {\it rpart} is {depicted in Figure \ref{tree}.}

Each leaf node corresponds to a sufficient condition for deceased status (True) or alive status (False).  There are three sufficient conditions for alive status, and two sufficient conditions for deceased status.  Tables \ref{cart} and  \ref{cartstats} contain the conditions and quantitative evaluations of the conditions produced by CART.  When comparing the results with those of \textsc{Conjecturing} in Tables \ref{alive}-\ref{deceasedstats} in the Appendix, we note that CART 
\begin{enumerate}
    \item produces many fewer conditions (3 versus 38 for alive status, 2 versus 40 for deceased status), 
    \item produces conditions that are in conjunctive normal form where each clause consists of a single numeric bound while \textsc{Conjecturing} tends to leverage nonlinear relationships among invariants as the basis for conditions, 
    \item produces two conditions with much larger support in the test data than those produced by \textsc{Conjecturing} (node 4 has 36,531 and node 7 has 12,443),  
    \item produces only two conditions that have lift greater than 1.0 (node 4 has lift 1.06 and node 7 has lift 2.85), and 
    \item does not produce conditions with better precision or lift than the best conditions produced by \textsc{Conjecturing}.  
\end{enumerate}
We note that both CART and \textsc{Conjecturing} are able to leverage categorical variables for conditions, though CART does not do so for this training set.  An example of such a condition is conjecture 27 in Table \ref{deceased} in the Appendix.   

Similar to many decision tree frameworks, CART leverages univariate bounds as component properties in its invariant clauses.  \textsc{Conjecturing} is unlikely to derive numeric bounds for individual features but instead produces more nonlinear relationships between invariants.  Decision tree frameworks such as CART and \textsc{Conjecturing} are complementary approaches for discovery of patterns among numeric and categorical features, but we see that \textsc{Conjecturing} is capable of producing more complex yet interpretable relationships.

\begin{table}[!ht]
\caption{\label{cart}Conditions from CART for Alive/Deceased Status Among Those with COVID}
\[
\begin{array}{lr}
\textrm{Node Number} & \textrm{Sufficient Condition} \\
\hline
\hline
4 & \textit{age} < 66.48 \ \& \ \textit{activeCarePlanLength} < 33.30 \rightarrow  \textit{Alive}\\
6 & \textit{age} \geq 66.48 \ \& \ \textit{meanPotassium} \geq 4.85 \rightarrow  \textit{Alive}\\
10 & \textit{age} < 66.48 \ \& \ \textit{activeCarePlanLength} \geq 33.30 \ \& \ \textit{bodyHeight} \geq 167.15 \rightarrow  \textit{Alive}\\
\hline
7 & \textit{age} \geq 66.48 \ \& \ \textit{meanPotassium} < 4.85 \rightarrow  \textit{Deceased}\\
11 & \textit{age} < 66.48 \ \& \ \textit{activeCarePlanLength} \geq 33.30 \ \& \ \textit{bodyHeight} < 167.15 \rightarrow  \textit{Deceased} \\ \hline
\end{array}
\]

\end{table}

\begin{table}[!ht]
\begin{center}
\caption{\label{cartstats}Evaluation of Conditions from CART Among Those with COVID}
\begin{tabular}{lrrrr}
Node Number & Consequent & Precision & Support & Lift \\
\hline
4  & \textit{Alive}   & 98.28\%   & 36531 & 1.06 \\
6  & \textit{Alive}   & 81.69\%   & 2070 & 0.88 \\
10 & \textit{Alive}   & 91.57\%   & 5371 & 0.99 \\
\hline
7  & \textit{Deceased}& 21.17\%   & 12443 & 2.85 \\
11 & \textit{Deceased}& 5.37\%    & 3746 & 0.72 \\ \hline
\end{tabular}

\end{center}
\end{table}

%\begin{table}[h!]
%\caption{\label{icu}Conjectures for ICU Status among those with COVID-19}
%\[
%\begin{array}{rcl} 
%\textit{healthcareCoverage}  < e^{(2\times 10^{\textit{medicationsLifetimePercCovered}})} & \rightarrow & \textit{no-ICU} \\
%\textit{healthcareExpenses}  > e^{-\textit{DALY}+\textit{QALY}} & \rightarrow & \textit{no-ICU} \\
%\textit{healthcareExpenses}  < proceduresLifetime^{e^{immunizationsLifetime}} & \rightarrow & \textit{no-ICU} \\
%\textit{healthcareCoverage}  > (\textit{encountersLifetimeTotalCost}-1)\times \textit{age} & \rightarrow & \textit{no-ICU} \\
%\textit{healthcareCoverage}  < \textit{encountersLifetimeTotalCost}\times \sqrt \textit{immunizationsLifetime} & \rightarrow & \textit{no-ICU} \\
%\textit{healthcareCoverage}  < e^{\sqrt{\textit{QALY}}+1} & \rightarrow & \textit{no-ICU} \\
%\textit{latitude}  < \textit{lifetimeConditionLength}/(\textit{medicationsLifetime}+1) & \rightarrow & \textit{no-ICU} \\
%\textit{hyperglycemiaDisorder}  & \rightarrow & \textit{ICU} \\
%\textit{coronaryHeartDisease}  & \rightarrow & \textit{ICU} \\
%\textit{medicationsLifetime}  > \max(\textit{plateletDistribution},e^{\textit{activeConditions}}) & \rightarrow & \textit{ICU} \\
%\textit{latitude}  > \max(\textit{QALY},1/\textit{deviceLifetimeLength}) & \rightarrow & \textit{ICU} \\ 
%\textit{formerSmoker}  & \rightarrow & \textit{ICU} \\
%\textit{lifetimeCarePlanLength}  > \textit{encountersLifetimePayerCoverage}/\sqrt{\textit{Latitude}} & \rightarrow & \textit{ICU} 
%\end{array}
%\] 
%\end{table}

\section{Conclusions}
\label{conclusions}
We have demonstrated that automated search for conjectured feature-relations can {support learning from data}.  The discovery of these kinds of feature relationships can also initiate new collaboration with domain scientists and lead to new scientific knowledge.

Our \textsc{Conjecturing} framework was able to recover the functional form for gravity with only the measured force, masses, and distance.  The framework also recovered a hidden interaction between price per square foot, square footage, and price in real estate data.  Using synthetic patient-level COVID-19 data, the framework produced conjectures that provide insight into the risk of death.  

The current version of \textsc{Conjecturing} requires that conjectures are true for every example.  Future research will {further investigate the ability of the framework to handle noisy data and investigate adjustments to} the algorithm to better handle noisy data such as generating conjectures that do not necessarily hold for all examples.  
%{\color{blue} In addition to noise for the invariant of interest, understanding the effect of noisy measurements for other invariants will be valuable as well.}  
If the handling of noise can be improved, then \textsc{Conjecturing} may be able to be adapted to support predictive modeling efforts.  

{As indicated by the motivating examples from graph theory, it can be desirable to retain multiple bounds or conditions produced by conjecturing because different conjectures might be better descriptions of system behavior for certain examples and not others.  In most of the experiments reported here, we evaluated conjectures as if they were each an entire model of the system and used standard ways of evaluating them.  Additional metrics besides those considered in this work might also be helpful in evaluating conjectures.  An interesting and important avenue of future research is how to better evaluate the ability of groups of conjectures to model system behavior.}

If the \textsc{Conjecturing} framework can provide functional relationships without constants of proportionality, the constant can be determined using regression with the original data.  Suppose that the \textsc{Conjecturing} framework indicates a relationship between the response $y$ and predictors $x$ of the form $y \leq b_1 f(x)$ for an unknown constant $b_1$.  A regression model can be fit of the form $\hat{y} = b_0 + b_1 f(x)$ using the data $(x_i, y_i)$, $i=1,\ldots,n$.  The best strategy for determining constants of proportionality is another avenue for future research.  { Constants can provide unit consistency to conjectures.  A method to search for unit consistent constants could facilitate the selection of meaningful conjectures.}

Another area for potential research involves the so-called $p >> n$ problem. That is, if the number of features is larger than the number of observations, then there are insufficient degrees of freedom to estimate a linear model with all $p$ features or any more complex model.  In such situations, feature and/or model selection tools are needed to search over potentially large model spaces. As desired model complexity increases (e.g. consideration of interaction terms), searching over such large model spaces can become computationally prohibitive. For instance, suppose an investigator seeks to identify a  model by selecting the ``best'' subset from among 10 features and their associated 45 two-way interactions. In this example, simply considering models with only 10 variables requires searching over a model space larger than 2.9 billion. Future research will investigate the ability of the \textsc{Conjecturing} framework to simplify model spaces and hence, provide a mechanism for a more expeditious search of plausible models.

% Appendix here
% Options are (1) APPENDIX (with or without general title) or
%             (2) APPENDICES (if it has more than one unrelated sections)
% Outcomment the appropriate case if necessary
%
% \begin{APPENDIX}{<Title of the Appendix>}
% \end{APPENDIX}
%
%   or
%
\newpage
\begin{APPENDICES}
\section{Additional Results for \textsc{Conjecturing} for Recovery of Equations}
\begin{table}[!ht]
{\footnotesize
\caption{\label{aif2}Additional Results for \textsc{Conjecturing} on Datasets from \citep{UdrescuTegmark20}}
\begin{tabular}{lllllll}
         &          & Number of  &            &         &           &   \\
Instance & Equation & Invariants & Complexity &  Noise  & Time (s)  & NRMSE  \\
\hline\hline
I.6.2.a & $f=e^{-\theta^2/2}/\sqrt{2\pi}$                            & 2 & 9  & $10^{-2}$  & 16    & 0.154 \\
I.6.2 & $f=e^{-\theta^2/2\sigma^2}/\sqrt{2\pi\sigma^2}$              & 3 & 13 & $10^{-4}$  & 2992  & 0.614 \\
I.6.2.b & $f=e^{-(\theta-\theta_1)^2/2\sigma^2}/\sqrt{2\pi\sigma^2}$ & 5 & 16 & $10^{-4}$  & 4792  & 1.188\\
I.8.14 & $d=\sqrt{(x_2-x_1)^2+(y_2-y_1)^2}$                          & 4 & 10 & $10^{-4}$  & 544   & 1.323\\
I.9.18 & $\frac{Gm_1m_2}{(x_2-x_1)^2+(y_2-y_1)^2+(z_2-z_1)^2}$       & 9 & 17 & $10^{-5}$  & 5975  & 1.252 \\
I.10.7 & $m=\frac{m_0}{\sqrt{1-\frac{v^2}{c^2}}}$                    & 3 & 9  & $10^{-4}$  & 14    & 0.034 \\
I.11.19 & $A=x_1y_1+x_2y_2+x_3y_3$                                   & 6 & 11 & $10^{-3}$  & 184   & 0.633\\
I.12.1 & $F=\mu N_n$                                                 & 2 & 3  & $10^{-3}$  & 12    & 0.002\\
I.12.2 & $F=\frac{q_1q_2}{4\pi\epsilon r^2}$                         & 5 & 12 & $10^{-2}$  & 17    & 0.885\\
I.12.4 & $E_f = \frac{q_1}{4\pi\epsilon r^2}$                        & 4 & 11 & $10^{-2}$  & 12    & 0.525 \\ \hline
\end{tabular}}

\end{table}
\newpage
\section{Features Generated for COVID-19 Data}
\begin{table} [!ht]
\caption{\label{covidfeatures}Feature Definitions for COVID-19 Data}
\begin{tabular}{l|p{4.0in}}
\textbf{Feature Name} & \textbf{Definition} \\ \hline
$healthcareExpenses$ & The total lifetime cost of healthcare to the patient          \\
$healthcareCoverage$ & The total lifetime cost of healthcare services that were covered by payers          \\
$latitude$ & Latitude of patient's home address \\
$longitude$ & Longitude of patient's home address\\
$age$ & Current age of patient\\
$numAllergies$ & Number of ongoing patient allergies\\
$activeCarePlans$ & Number of current care plans\\
$lifetimeCarePlans$ & Number of lifetime care plans\\
$activeCarePlanLength$ & Length of time under current care plans\\
$lifetimeCarePlanLength$ & Total lifetime length under care plans\\
$activeConditions$ & Number of current health conditions\\
$lifetimeConditions$ & Number of lifetime health conditions\\
$activeConditionLength$ & Amount of time since current health condition(s) diagnosis\\
$lifetimeConditionLength$ & Amount of time since first diagnosis of a health condition \\
$deviceLifetimeLength$ & Total length of time using a medical device (e.g. pacemaker)  \\
$encountersCount$ & Total number of encounters with a healthcare professional\\
$encountersLifetimeTotalCost$ & Total lifetime cost of healthcare encounters \\
$encountersLifetimeBaseCost$ & Total lifetime cost of healthcare encounters, not including any line item costs related to medications, immunizations, procedures, or other services\\
$encountersLifetimePayerCoverage$ & Total lifetime cost of healthcare encounters that were covered by payers \\
$encountersLifetimePercCovered$ & Percentage of lifetime cost of healthcare encounters that were covered by payer\\
$imagingStudiesLifetime$ & Number of lifetime imaging diagnostics (e.g. MRI) performed on patient\\
$immunizationsLifetime$ & Number of lifetime immunizations received by patient \\
$immunizationsLifetimeCost$ & Total lifetime cost of all immunizations received by patient\\
$medicationsLifetime$ & Number of lifetime medications prescribed \\
$medicationsLifetimeCost$ & Total lifetime cost of medications\\
$medicationsLifetimePercCovered$ & Percentage of lifetime medication cost coverered by payer\\
$medicationsLifetimeLength$ & Total lifetime length on prescribed medications\\
$medicationsLifetimeDispenses$ & Total lifetime number of prescription dispenses\\
$medicationsActive$ & Number of current prescriptions \\
$proceduresLifetime$ & Number of lifetime medical procedures (e.g. surgery) performed on patient\\
$proceduresLifetimeCost$ & Total lifetime cost of all medical procedures performed on patient\\
%$QOLS$ & The average Quality of Life Scores for all patients enrolled with payer \\
%$QALY$ & Quality Adjusted Life Years\\
%$DALY$ & Disability Adjusted Life Years\\
\hline
\end{tabular}
\end{table}
\newpage

\section{Results for \textsc{Conjecturing} Applied to COVID Data}

\begin{table}[!ht]
\caption{\label{alive}Conjectures for Alive Status Among Those with COVID}
{\footnotesize
\[
 \begin{array}{lr} 
& \textrm{Conjecture}  \\
 \hline\hline
 1&  \textit{activeConditionLength} > \textit{age}^2/\textit{latitude} \rightarrow \textit{Alive}  \\ 
 2&  \textit{medicationsLifetime} < -\textit{immunizationsLifetimeCost} + 2 \times \textit{proceduresLifetime} \rightarrow \textit{Alive}  \\ 
 3&  \textit{medicationsActive} < \min\{\textit{sodium},\lfloor \textit{QOLS}\rfloor\} \rightarrow \textit{Alive}  \\ 
 4&  \textit{activeCarePlans} < e^{\textit{medicationsLifetimePercCovered}} - \textit{immunizationsLifetime} \rightarrow \textit{Alive}  \\ 
 5&  \textit{activeCarePlanLength} < 10^\textit{encountersLifetimePercCovered} \times \textit{activeCarePlans} \rightarrow \textit{Alive}  \\ 
 6&  \textit{lifetimeConditionLength} > \sqrt{\textit{QALY}}^\textit{activeConditions} \rightarrow \textit{Alive}  \\ 
 7&  \textit{lifetimeCarePlans} > \textit{encountersCount}/2  \rightarrow \textit{Alive}  \\ 
 8 & \textit{lifetimeCarePlans} > \min\{\textit{triglycerides}, \lfloor \textit{lifetimeConditions}\rfloor\} \rightarrow \textit{Alive}  \\ 
 9&  \textit{encountersCount} > \textit{activeConditions} + \textit{medicationsLifetimeCost} + 1 \rightarrow \textit{Alive}  \\ 
 10&  \textit{diabetes } \& \textit{ latitude} < \sqrt{\textit{totalCholesterol}} + \textit{meanCarbonDioxide} \rightarrow \textit{Alive}  \\ 
 11&  \textit{activeCarePlans} > \textit{medicationsLifetime}^\textit{medicationsLifetimeCost}  \rightarrow \textit{Alive}  \\ 
 12&  \textit{lifetimeCarePlans} > \sqrt{\textit{enountersCount}} +  \textit{QOLS} \rightarrow \textit{Alive}  \\ 
 13&  \textit{age} < \textit{lifetimeConditions} \times \log(\textit{latitude})  \rightarrow \textit{Alive}  \\ 
 14&  \textit{anemiaDisorder } \& \textit{ activeConditionLength} < \min \{\textit{ureaNitrogen}, |\textit{activeCarePlanLength}|\} \rightarrow \textit{Alive}  \\ 
 15&  \textit{lifetimeCarePlans} > \max\{ \textit{DALY}, \lfloor \textit{potassium}\rfloor\} \rightarrow \textit{Alive}  \\ 
 16&  \textit{healthcareCoverage} > \lfloor \textit{DALY}\rfloor \times \textit{lifetimeCarePlanLength} \rightarrow \textit{Alive}  \\ 
 17&  \textit{lifetimeConditionLength} > (\textit{encountersLifetimeTotalCost} - 1)/\textit{proceduresLifetime} \rightarrow \textit{Alive}  \\ 
 18&  \textit{healthcareCoverage} > \textit{lifetimeConditionLength}^2/\textit{imagingStudiesLifetime}  \rightarrow \textit{Alive}  \\ 
 19& \textit{immunizationsLifetimeCost} > (\textit{bodyHeight}-1)^{\textit{immunizationsLifetime}} \rightarrow \textit{Alive}  \\ 
 20&  \textit{numAllergies} < \textit{activeCarePlanLength} - \textit{age} + 1  \rightarrow \textit{Alive}  \\ 
 21& \textit{activeCarePlans} < \lfloor \textit{microalbuminCreatineRatio} \rfloor - \textit{proceduresLifetimeCost} \rightarrow \textit{Alive}  \\ 
 22&  \textit{latitude} < \sqrt{\textit{encountersLifetimeTotalCost}} - \textit{medicationsLifetime} \rightarrow \textit{Alive}  \\ 
 23&  \textit{bodyMassIndex40} \rightarrow \textit{Alive}  \\ 
 24&  \textit{activeCarePlanLength} > \textit{age} + \textit{proceduresLifetime} -1 \rightarrow \textit{Alive}  \\ 
 25&  \textit{medicationsLifetimeLength} < 2\times \textit{activeCarePlan}\times \textit{deviceLifetimeLength}  \rightarrow \textit{Alive}  \\ 
 26&  \textit{carbonDioxide} > \textit{respiratoryRate}\times \lfloor \textit{hemoglobinA1cHemoglobinTotalInBlood}\rfloor \rightarrow \textit{Alive}  \\ 
 27&  \textit{osteoporosisDisorder } \& \textit{ lifetimeCarePlanLength} < \min\{\textit{painSeverity},2\times\textit{activeCarePlanLength}\}  \rightarrow \textit{Alive}  \\ 
 28&  \textit{healthcareCoverage} < \sqrt{ \textit{encountersLifetimePayerCoverage}} \times \textit{medicationsLifetime} \rightarrow \textit{Alive}  \\ 
 29&  \textit{respiratoryRate} < \textit{painSeverity} + \lceil \textit{leukocytesVolumeInBlood}\rceil  \rightarrow \textit{Alive}  \\ 
 30&  \textit{meanPainSeverity} > \max\{\textit{proceduresLifetime}, \lfloor \textit{hemoglobinA1cHemoglobinTotalInBlood}\rfloor\} \rightarrow \textit{Alive}  \\ 
 31&  \textit{prediabetes } \& \textit{ meanDiastolicBloodPressure} > \lfloor \textit{carbonDioxide}\rfloor + \textit{meanHeartRate} \rightarrow \textit{Alive}  \\
 32&  \textit{latitude} < \textit{encountersCount}\times \lfloor\textit{QOLS}\rfloor \rightarrow \textit{Alive}  \\ 
 33&  \textit{painSeverity} < \lfloor \textit{meanPainSeverity}\rfloor - 1 \rightarrow \textit{Alive}  \\ 
 34& \textit{lifetimeCarePlans} > \max\{\textit{DALY}, \lfloor \textit{potassium} \rfloor\} \rightarrow \textit{Alive}\\ 
 35&  \textit{latitude} < 2\times \textit{DALY}\times \textit{encountersLifetimePercCovered} \rightarrow \textit{Alive}  \\ 
 36&  \textit{activeCarePlanLength} > \max \{\textit{sodium}, \lceil \textit{hematocritVolume}\rceil\} \rightarrow \textit{Alive}  \\ 
 37&  \textit{medicationsLifetimePercCovered} > \textit{latitude}^2/\textit{medicationsLifetimeDispense} \rightarrow \textit{Alive}  \\ 
 38 & \textit{healthcareCoverage} > \frac{\textit{healthcareExpenses}}{2\times \textit{encountersLifetimePercCovered}} \rightarrow \textit{Alive}  \\ \hline
 \end{array} 
 \]
}
 \end{table}
 
 \newpage
 
 \begin{table}[!ht]
 \caption{\label{alivestats}Evaluation of Conjectures for Alive Status Among Those with COVID}
 \centering
 \begin{tabular}{rrrr}
Conjecture & Precision & Support & Lift\\
\hline
1 & 99.11\% & 3042 & 1.07\\
2 & 98.54\% & 3700 & 1.06\\
3 & 98.51\% & 9480 & 1.06\\
4 & 98.25\% & 8231 & 1.06\\
5 & 98.23\% & 6103 & 1.06\\
6 & 98.05\% & 7937 & 1.06\\
7 & 97.79\% & 4161 & 1.06\\
8 & 97.65\% & 5453 & 1.05\\
9 & 97.65\% & 2939 & 1.05\\
10 & 97.41\% & 1969 & 1.05\\
11 & 97.07\% & 4774 & 1.05\\
12 & 96.79\% & 2242 & 1.05\\
13 & 96.61\% & 1650 & 1.04\\
14 & 95.65\% & 2045 & 1.03\\
15 & 95.39\% & 6311 & 1.03\\
16 & 95.31\% & 1236 & 1.03\\
17 & 95.20\% & 1687 & 1.03\\
18 & 95.07\% & 4017 & 1.03\\
19 & 94.32\% & 440 & 1.02\\
20 & 94.00\% & 500 & 1.02\\
21 & 93.59\% & 1498 & 1.01\\
22 & 93.25\% & 1185 & 1.01\\
23 & 92.09\% & 834 & 0.99\\
24 & 91.01\% & 1213 & 0.98\\
25 & 90.79\% & 999 & 0.98\\
26 & 89.77\% & 831 & 0.97\\
27 & 89.41\% & 727 & 0.97\\
28 & 88.61\% & 2389 & 0.96\\
29 & 88.27\% & 358 & 0.95\\
30 & 88.21\% & 704 & 0.95\\
31 & 87.67\% & 1890 & 0.95\\
32 & 87.20\% & 2250 & 0.94\\
33 & 87.14\% & 583 & 0.94\\
34 & 85.79\% & 1612 & 0.93\\
35 & 84.50\% & 755 & 0.91\\
36 & 82.76\% & 586 & 0.89\\
37 & 76.22\% & 677 & 0.82\\
38 & 74.75\% & 99 & 0.81 \\ \hline
\end{tabular}
\end{table}
 
 \newpage

\begin{table}[!ht]
\caption{\label{deceased}Conjectures for Deceased Status Among Those with COVID}
{\footnotesize
\[
 \begin{array}{lr} 
\textrm{Conjecture} & \textrm{Sufficient Condition}  \\
 \hline\hline
1 & \textit{longitude} > -\textit{age} \times \textit{medicationsLifetimePercCovered}  \rightarrow \textit{Deceased} \\
2 & \textit{medicationsActive} > \lfloor \textit{hemoglobinA1cHemoglobinTotalInBlood} \rfloor \rightarrow \textit{Deceased} \\
3 & \textit{age} > \textit{carbonDioxide}\times \lfloor \textit{potassium} \rfloor \rightarrow \textit{Deceased} \\
4 & \textit{deviceLifetimeLength} \leq 2\times \textit{creatinine}^\textit{healthcareExpenses} \rightarrow \textit{Deceased} \\
5 & \textit{implantableCardiacPacem}  \rightarrow \textit{Deceased} \\
6 & \textit{latitude} < \log(\textit{age})/\log(10)^\textit{activeCarePlans}  \rightarrow \textit{Deceased} \\
7 & \textit{medicationsActive} > \lceil \log(\textit{alkalinePhosphataseEnzymatic Activity})/\log (10) \rceil \rightarrow \textit{Deceased} \\
8 & \textit{immunizationsLifetimeCost} < \textit{age} \times \textit{immunizationsLifetime}^2 \rightarrow \textit{Deceased} \\
9 & \textit{colonoscopy } \& \textit{ coronaryHeartDisease} \rightarrow \textit{Deceased} \\
10 & \textit{activeCarePlans} < \min \{ \textit{deviceLifetimeLength}, \textit{medicationsActive}\}  \rightarrow \textit{Deceased} \\
11 & \textit{glucose} > \lceil \textit{creatinine} \times \textit{meanGlucose} \rightarrow \textit{Deceased} \\
12 & \textit{bodyWeight} > \lfloor \textit{meanBodyWeight} \rfloor + 1 \rightarrow \textit{Deceased} \\
13 & \textit{healthcareExpenses} < \textit{deviceLifetimeLength}^2 \times \textit{lifetimeConditionLength} \rightarrow \textit{Deceased} \\
14 & \textit{lifetimeConditions} > \textit{activeCarePlans} + \lfloor \textit{ureaNitrogen} \rfloor \rightarrow \textit{Deceased} \\
15 & \textit{activeCarePlans} > \lfloor \log(\textit{triglycerides})\rfloor  \rightarrow \textit{Deceased} \\
16 & \textit{healthcareExpenses} < \textit{lifetimeConditionLength}^2  + \textit{encountersLifetimeTotalCost} \rightarrow \textit{Deceased} \\
17 & \textit{overlappingMalignantNeo} \rightarrow \textit{Deceased}\\ 
18 & \textit{latitude} > \textit{ureaNitrogen} \lceil \textit{albuminMassVolumeInSerumOrPlasma}\rceil  \rightarrow \textit{Deceased} \\
19 &  \textit{activeConditionLength} > \textit{erythrocytesVolumeInBlood} \times \lceil \textit{hemoglobinMassVolumeInBlood} \rceil \rightarrow \textit{Deceased} \\
20 & \textit{chronicObstructiveBronc} \rightarrow \textit{Deceased} \\
21 & \textit{longitude} > \sqrt{\textit{healthcareCoverage}} - \textit{encountersCount} \rightarrow \textit{Deceased} \\
22 & \textit{age} > 10^\textit{medicationsActive} - \textit{longitude} \rightarrow \textit{Deceased} \\
23 & \textit{chloride} < \lfloor \textit{meanChloride}\rfloor - \textit{lifetimeCarePlans} \rightarrow \textit{Deceased} \\
24 & \textit{medicationsActive} > \max\{\textit{respiratoryRate}, \log(\textit{latitude})\} \rightarrow \textit{Deceased} \\
25 & \textit{localizedPrimaryOsteoa}  \rightarrow \textit{Deceased} \\
26 & \textit{rheumatoidArthritis} \rightarrow \textit{Deceased} \\
27 & \textit{chronicPain } \& \textit{ smokesTobaccoDaily} \rightarrow \textit{Deceased} \\
28 & \textit{latitude} < \lfloor\textit{erythrocyteDistributionWidth}\rfloor - \textit{meanPainSeverity}   \rightarrow \textit{Deceased} \\
29 & \textit{activeCarePlans} > 10^\textit{medicationsActive}/ \textit{imagingStudiesLifetime}  \rightarrow \textit{Deceased} \\
30 &\textit{tubalPregnancy}  \rightarrow \textit{Deceased} \\
31 & \textit{activeConditions} < \textit{medicationsActive}^2- \textit{medicationsLifetime} \rightarrow \textit{Deceased} \\
32 & \textit{alcoholism } \& \textit{ majorDepressionDisorder} \rightarrow \textit{Deceased} \\
33 & \textit{creatinine} < \lceil \textit{meanCreatinine} \rceil/\textit{lifetimeCarePlans} \rightarrow \textit{Deceased} \\
34 & \textit{healthcareCoverage} < \textit{encoutnersLifetimePayerCoverage}\times \log (\textit{latitude})/\log (10)  \rightarrow \textit{Deceased} \\
35 & \textit{encountersCount} < \min \{ \textit{DALY}, 10^\textit{immunizationsLifetime}\}  \rightarrow \textit{Deceased} \\
36 & \textit{lifetimeCarePlanLength} > \textit{age} +  e^{\textit{medicationsLifetime}} \rightarrow \textit{Deceased} \\
37 & \textit{healthcareCoverage} < \textit{activeCondionLength}^2- \textit{encountersLifetimeTotalCost} \rightarrow \textit{Deceased} \\
38 & \textit{age} > 1/2 \times \textit{healthcareExpenses}/\textit{immunizationsLifetimeCost} \rightarrow \textit{Deceased} \\
39 & \textit{activeCarePlanLength} > \textit{activeConditionLength}\times e^{\textit{DALY}} \rightarrow \textit{Deceased} \\
40 & \textit{medicationsLifetime} < \sqrt{\textit{encountersLifetimePayerCoverage}} - \textit{age} \rightarrow \textit{Deceased} \\ \hline
\end{array} 
\]
}
\end{table}
\newpage

 \begin{table}[!ht]
 \caption{\label{deceasedstats}Evaluation of Conjectures for Deceased Status Among Those with COVID}
 \centering
 \begin{tabular}{rrrr}
Conjecture & Precision & Support & Lift\\
\hline
1 & 30.91\% & 372 & 4.15\\
2 & 24.44\% & 2954 & 3.29\\
3 & 23.22\% & 1722 & 3.12\\
4 & 22.47\% & 632 & 3.02\\
5 & 22.39\% & 844 & 3.01\\
6 & 21.74\% & 1490 & 2.92\\
7 & 21.44\% & 4427 & 2.88\\
8 & 21.24\% & 2199 & 2.85\\
9 & 21.01\% & 257 & 2.82\\
10 & 20.90\% & 799 & 2.81\\
11 & 20.03\% & 1058 & 2.69\\
12 & 18.25\% & 548 & 2.45\\
13 & 17.56\% & 467 & 2.36\\
14 & 17.54\% & 1898 & 2.36\\
15 & 17.31\% & 3328 & 2.33\\
16 & 17.23\% & 940 & 2.32\\
17 & 16.92\% & 130 & 2.27\\
18 & 16.13\% & 1091 & 2.17\\
19 & 15.68\% & 797 & 2.11\\
20 & 14.11\% & 900 & 1.90\\
21 & 14.08\% & 781 & 1.89\\
22 & 14.08\% & 2230 & 1.89\\
23 & 11.99\% & 884 & 1.61\\
24 & 11.84\% & 1884 & 1.59\\
25 & 11.76\% & 2159 & 1.58\\
26 & 11.44\% & 201 & 1.54\\
27 & 10.54\% & 607 & 1.42\\
28 & 10.27\% & 1724 & 1.38\\
29 & 9.39\% & 213 & 1.26\\
30 & 8.85\% & 2147 & 1.19\\
31 & 8.22\% & 4150 & 1.10\\
32 & 7.79\% & 1129 & 1.05\\
33 & 7.77\% & 2408 & 1.04\\
34 & 7.73\% & 634 & 1.04\\
35 & 5.20\% & 3209 & 0.70\\
36 & 3.89\% & 4602 & 0.52\\
37 & 3.85\% & 467 & 0.52\\
38 & 2.65\% & 339 & 0.36\\
39 & 2.29\% & 3188 & 0.31\\
40 & 1.30\% & 1001 & 0.17 \\ \hline
\end{tabular}
\end{table}

 \end{APPENDICES}

\newpage

% Acknowledgments here
\ACKNOWLEDGMENT{High Performance Computing resources provided by the High Performance Research Computing (HPRC) Core Facility at Virginia Commonwealth University (\url{https://chipc.vcu.edu}) were used for conducting the research reported in this work.} 

% References here (outcomment the appropriate case)

% CASE 1: BiBTeX used to constantly update the references
%   (while the paper is being written).
\bibliographystyle{informs2014} % outcomment this and next line in Case 1
\bibliography{larson} % if more than one, comma separated

% CASE 2: BiBTeX used to generate mypaper.bbl (to be further fine tuned)
%\input{mypaper.bbl} % outcomment this line in Case 2

%If you don't use BiBTex, you can manually itemize references as shown below.

%%%%%%%%%%%%%%%%%
\end{document}